\RequirePackage{fix-cm}
\RequirePackage{fixltx2e}
\documentclass[twoside,english,aip, jcp,floatfix]{revtex4-1}
\usepackage{mathpazo}

\usepackage[T1]{fontenc}
\usepackage[latin9]{inputenc}
\usepackage[a4paper]{geometry}
\usepackage{xcolor}
\usepackage{babel}
\usepackage{rotating}
\usepackage{amsmath}
\usepackage{amssymb}
\usepackage{stmaryrd}
\usepackage{graphicx}
\usepackage[unicode=true,pdfusetitle,
 bookmarks=false,
 breaklinks=true,pdfborder={0 0 0},pdfborderstyle={},backref=false,colorlinks=true]
 {hyperref}
\hypersetup{
 citecolor = blue, linkcolor = blue,  urlcolor  = blue}

\makeatletter

\providecommand{\tabularnewline}{\\}

\usepackage{orcidlink}

\newcommand{\beginappendix}{%
        \setcounter{table}{0}
        \renewcommand{\thetable}{A\arabic{table}}%
        \setcounter{figure}{0}
        \renewcommand{\thefigure}{A\arabic{figure}}%
     }

\makeatother

\begin{document}

\title{{\Large{}Validation of ML-UQ calibration statistics using simulated
reference values: a sensitivity analysis }}

\author{Pascal PERNOT \orcidlink{0000-0001-8586-6222}}

\affiliation{Institut de Chimie Physique, UMR8000 CNRS,~\\
Université Paris-Saclay, 91405 Orsay, France}
\email{pascal.pernot@cnrs.fr}

\selectlanguage{english}%
\begin{abstract}
\noindent Some popular Machine Learning Uncertainty Quantification
(ML-UQ) calibration statistics do not have predefined reference values
and are mostly used in comparative studies. In consequence, calibration
is almost never validated and the diagnostic is left to the appreciation
of the reader. Simulated reference values, based on synthetic calibrated
datasets derived from actual uncertainties, have been proposed to
palliate this problem. As the generative probability distribution
for the simulation of synthetic errors is often not constrained, the
sensitivity of simulated reference values to the choice of generative
distribution might be problematic, shedding a doubt on the calibration
diagnostic. This study explores various facets of this problem, and
shows that some statistics are excessively sensitive to the choice
of generative distribution to be used for validation when the generative
distribution is unknown. This is the case, for instance, of the correlation
coefficient between absolute errors and uncertainties (CC) and of
the expected normalized calibration error (ENCE). A robust validation
workflow to deal with simulated reference values is proposed.
\end{abstract}
\maketitle

\section{Introduction}

Most calibration statistics are used to compare UQ methods or datasets
but are missing a reference value for validation. For instance, the
correlation coefficient between absolute errors and uncertainties\citep{Tynes2021}
should be positive, but has no predefined reference value to compare
with\citep{Pernot2022b}. This is also the case for some conditional
calibration statistics, such as the expected normalized calibration
error (ENCE)\citep{Levi2022}, which value depends moreover on the
binning scheme\citep{Pernot2023a_arXiv}.

The use of \emph{probabilistic}\citep{Pernot2022c} or \emph{simulated}\citep{Rasmussen2023}
reference values has been recently proposed to palliate the absence
of reference for a calibration statistic. A simulated reference value
is estimated by applying the statistic to synthetic datasets containing
the actual uncertainties and simulated errors generated from the uncertainties
using a model generative distribution $D$. 

The main issue with simulated references is the choice of $D$. Some
calibration statistics depend explicitly on $D$ (typically a normal
distribution): calibration curves and calibration error\citep{Tran2020},
negative log-likelihood (NLL)... This scenario fixes the choice of
$D$ to simulate a reference value. However, this is not the case
for many other calibration statistics, such as the calibration error,
confidence curves\citep{Pernot2022c} or reliability diagrams/ENCE\citep{Pernot2023a_arXiv}.
Without constraint on $D$, the question arises of the sensitivity
of such simulated reference values to the choice of generative distribution.

The case of confidence curves\citep{Pernot2022c} enables to underline
the main targets of the present study. A confidence curve is obtained
by estimating an error statistic (e.g. the mean absolute error, MAE)
for a dataset iteratively pruned from its largest uncertainties. Plotted
against the percentage of pruned data, the confidence curve tests
how the largest errors are associated with the largest uncertainties.
Ideally, a confidence curve should be monotonously decreasing, but
there is no predefined reference curve to assess its quality (the
so-called \emph{oracle} is unreachable in ML regression tasks\citep{Pernot2022c}).
In this context, a confidence curve can inform us on the quality of
the link between errors and uncertainties, but not on calibration.
Using a simulated reference curve can solve this problem, at the cost
of a choice for the generative distribution $D$, about which Pernot\citep{Pernot2022c}
has raised two main points: 
\begin{itemize}
\item the sensitivity of the simulated reference curve to $D$ depends on
the error statistic used to build the confidence curve: the MAE is
very sensitive to $D$, while the root mean-squared error (RMSE) is
not;
\item for validation, a confidence band can be estimated from simulated
reference curves. For all error statistics, the width of the confidence
band depends on $D$, leading to ambiguous validation diagnostics
when $D$ is not constrained. 
\end{itemize}
The present study considers these two points for other calibration
statistics. More specifically, it aims to test (i) how simulated reference
values for calibration statistics are sensitive to the choice of a
generative distribution, (ii) how this impacts the validation procedure,
and (iii) whether the uncertainty on the simulated reference value
is a good choice of metric for validation. The focus is limited here
to statistics linked to ML regression tasks: the correlation coefficient
between absolute errors and uncertainties (CC), an average calibration
statistic (the mean squared \emph{z}-scores, ZMS), and two conditional
calibration statistics (the ENCE, its ZMS-derived analog ZMSE).\textcolor{red}{{} }

The next section defines the calibration statistics, the validation
approach, and proposes a workflow dealing with the main issues of
simulated references. Sect.\,\ref{sec:Applications} presents the
application of these methods to an ensemble of nine datasets issued
from the ML-UQ literature. The article proceeds with a discussion
of the main findings (Sect.\,\ref{sec:Discussion}), and the conclusions
are reported in Sect.\,\ref{sec:Conclusion}.

\section{Validation scores for calibration and correlation statistics\label{sec:Validation-methods-for}}

This section presents the simulation method for the estimation of
calibration statistics references for commonly used correlation, \emph{average}
calibration and \emph{conditional} calibration statistics in order
to define adequate methods for their validation. 

\subsection{Probabilistic generative model\label{subsec:Probabilistic-reference-1}}

Let us consider a dataset composed of \emph{paired} errors and uncertainties
$\left\{ E_{i},u_{E_{i}}\right\} _{i=1}^{M}$ to be tested for calibration.
A few variance-based UQ validation statistics and methods, based on
the correlation between the absolute errors and uncertainties (rank
correlation coefficient (Sect.\,\ref{subsec:Some-calibration-statistics})
and confidence curve), avoid the need of a probabilistic model linking
those variables\citep{Pernot2022b}. In contrast, most of the variance-based
UQ calibration statistics are built implicitly on a probabilistic
model
\begin{equation}
E_{i}\sim D(\mu=0,\sigma=u_{E_{i}})\label{eq:probmod}
\end{equation}
or, equivalently,
\begin{equation}
E_{i}=u_{E_{i}}\varepsilon_{i}\label{eq:probmod-1}
\end{equation}
linking errors to uncertainties, where $\varepsilon_{i}$ is a random
number with zero-centered and unit-variance probability density function
{[}$\varepsilon\sim D(\mu=0,\sigma=1)${]}. This model states that
errors should be unbiased ($\mu=0$) and that uncertainty quantifies
the \emph{dispersion} of errors, according to the metrological definition.\citep{GUM} 

Except in some instances where $D$ is constrained by the method used
to generate the dataset (gaussian process, normal likelihood models...),
the shape of $D$ is unknown. This is notably the case when uncertainties
are obtained\emph{ by post-hoc }calibration\citep{Levi2022,Busk2022}. 

Note that $D$ should not be mistaken for the distribution of errors,
which is a \emph{compound distribution} between $D$ and the distribution
of uncertainties. Let us assume that the errors are drawn from a distribution
$D$ (Eq.\,\ref{eq:probmod}) with a scale parameter $\sigma$, itself
distributed according to a distribution $G$. The distribution of
errors, H, is then a \emph{scale mixture distribution} with probability
density function 
\begin{equation}
p_{H}(E)={\displaystyle \int_{0}^{\infty}p_{D}(E|\sigma)\,p_{G}(\sigma)\thinspace d\sigma}
\end{equation}
\medskip{}

\noindent \textbf{Example - the NIG distribution.} For a normal distribution
$D=N(0,\sigma)$ and a distribution of variances described by an inverse
gamma distribution $\sigma^{2}\sim\Gamma^{-1}(\frac{\nu}{2},\frac{\nu}{2})$,
the compound distribution $H$ is a Student's-\emph{t} distribution
with $\nu$ degrees of freedom, noted $t(\nu)$. This is a special
case of the so-called Normal Inverse Gamma (NIG) distribution used,
for instance, in evidential inference\citep{Amini2019}.\hfill{}
$\oblong$

When $D$ is unknown, there is no evidence that it should be uniform
across data space. To avoid unlimited complexity, this hypothesis
will be made in the following, without affecting the main conclusions
of the study.

\subsection{Calibration statistics derived from the generative model\label{subsec:Some-calibration-statistics}}

The generative model described above can be used to derive two families
of calibration statistics.

\subsubsection{The calibration error and related statistics.\label{par:RCE}}

The variance of the compound distribution of errors is obtained by
the \emph{law of total variance}, i.e.
\begin{align}
\mathrm{Var}_{H}(E) & =\left\langle \mathrm{Var}_{D}(E|\sigma)\right\rangle _{G}+\mathrm{Var}_{G}\left(\left\langle E|\sigma\right\rangle _{D}\right)\label{eq:totalVar-1}\\
 & =<u_{E}^{2}>+\mathrm{Var}_{G}\left(\left\langle E|\sigma\right\rangle _{D}\right)
\end{align}
where the first RHS term of Eq.\,\ref{eq:totalVar-1} has been identified
as the mean squared uncertainty $<u_{E}^{2}>$. This expression can
be compared to the standard expression for the variance
\begin{align}
\mathrm{Var}(E) & =<E^{2}>-<E>^{2}\label{eq:totalVar-1-1}
\end{align}
For an unbiased error distribution, one gets $\mathrm{Var}_{G}\left(\left\langle E|\sigma\right\rangle _{D}\right)=0$
and $<E>=0$, leading to
\begin{equation}
<E^{2}>=<u_{E}^{2}>\label{eq:RMS=00003DRMV}
\end{equation}
on which some popular calibration statistics are based. \medskip{}

\noindent \textbf{Example, followed.} Considering the NIG model, one
can easily verify from the properties of the Student's-\emph{t} distribution
that
\begin{equation}
<E^{2}>=\sigma^{2}\nu/(\nu-2)
\end{equation}
(using $<E>=0$), and from the Inverse Gamma distribution\citep{Evans2000}
that 
\begin{equation}
<u_{E}^{2}>=\sigma^{2}\nu/(\nu-2)
\end{equation}
so that Eq.\,\ref{eq:RMS=00003DRMV} is theoretically fulfilled.\hfill{}
$\oblong$

\medskip{}

Based on Eq.\,\ref{eq:RMS=00003DRMV}, the Relative Calibration Error
is defined as
\begin{equation}
RCE=\frac{RMV-RMSE}{RMV}\label{eq:RCE}
\end{equation}
where $RMSE$ is the root mean squared error $\sqrt{<E^{2}>}$ and
$RMV$ is the root mean variance ($\sqrt{<u_{E}^{2}>}$). The RCE
has been shown to be very sensitive to the presence of heavy tails
in the uncertainty and error distributions and to be unreliable for
a large portion of the studied ML-UQ datasets\citep{Pernot2024_arXiv}.
It is therefore not considered directly here, but it is used in a
bin-based statistic of \emph{conditional} calibration, the Expected
Normalized Calibration Error\citep{Levi2022}
\begin{equation}
ENCE=\frac{1}{N}\sum_{i=1}^{N}|RCE_{i}|
\end{equation}
where $RCE_{i}$ is estimated over the data within bin $i$. According
to the binning variable, the ENCE might be used to test \emph{consistency}
(binning on $u_{E}$) or \emph{adaptivity} (binning on input features).\citep{Pernot2023d} 

Pernot\citep{Pernot2023a_arXiv} has shown that the ENCE reference
value is not zero, and that it depends on the binning scheme. As demonstrated
in Appendix\,\ref{sec:Reference-values-for}, the ENCE depends on
the bins size and does not have a predefined reference value.

In case of heavy-tailed uncertainty distributions, the ENCE is not
expected to suffer from the same problem as the RCE when the binning
variable is the uncertainty\citep{Pernot2024_arXiv}. However, it
shares the same sensitivity in the case of heavy-tailed error distributions
or when the binning variable is not the uncertainty. 

\subsubsection{The ZMS and related statistics.\label{par:ZMS}}

Another approach to calibration based on Eq.\,\ref{eq:probmod} uses
scaled errors or \emph{z}-scores
\begin{equation}
Z_{i}=\frac{E_{i}}{u_{E_{i}}}\sim D(0,1)
\end{equation}
with the property
\begin{equation}
Var(Z)=1
\end{equation}
assessing average calibration for unbiased errors\citep{Pernot2022a,Pernot2022b}.
For biased errors, the calibration equation becomes
\begin{equation}
ZMS=<Z^{2}>=1\label{eq:ZMS}
\end{equation}
which is the preferred form for testing\citep{Pernot2023d} (ZMS stands
for \emph{z}-score's mean squares). This choice is motivated by the
use of the ZMS for binned data, where, even for an unbiased dataset,
one should not expect every bin to be unbiased. The ZMS does not depend
on a distribution hypothesis and its target value is 1. 

The negative log-likelihood (NLL) score can be written as\citep{Busk2023}
\begin{align}
NLL & =\frac{1}{2}\left(<Z^{2}>+<\ln u_{E}^{2}>+\ln2\pi\right)\label{eq:NLLdef}
\end{align}
It combines an \emph{average calibration} term, the ZMS,\citep{Zhang2023}
and a \emph{sharpness} term driving the uncertainties towards small
values\citep{Gneiting2007a} when the NLL is used as a loss function,
hence preventing the minimization of $<Z^{2}>$ by arbitrary large
uncertainties. The NLL is the logarithm of a normal probability density
function and therefore should be used only when the errors and uncertainties
are linked by a standard normal generative distribution ($D=N(0,1)$).
Knowing the reference value of $<Z^{2}>$, one can assign a reference
value to the NLL
\begin{equation}
NLL_{ref}=\frac{1}{2}\left(1+<\ln u_{E}^{2}>+\ln2\pi\right)\label{NLLref}
\end{equation}
Note that Rasmussen \emph{et al.}\citep{Rasmussen2023} treat the
NLL as if it had no predefined reference value and needed a simulated
reference. This is not the case when the NLL is defined by Eq.\,\ref{eq:NLLdef},
but it might be for other likelihood definitions. In the present case,
for a given set of uncertainties, validation of the NLL is equivalent
to the validation of the ZMS, with the additional constraint of a
normal generative model. 

By analogy with the ENCE, one can define a ZMS-based mean calibration
error\citep{Pernot2023c_arXiv}
\begin{equation}
ZMSE=\frac{1}{N}\sum_{i=1}^{N}|\ln ZMS_{i}|
\end{equation}
where $i$ runs over the $N$ bins and $ZMS_{i}$ is estimated with
the data in bin $i$. The logarithm accounts for the fact that ZMS
is a scale statistic (a ZMS of 2 entails a deviation of the same amplitude
as a ZMS of 0.5). As for the ENCE, the ZMSE measures conditional calibration,
i.e. consistency if the binning variable is the uncertainty or adaptivity
if it is an input feature. Also, the ZMSE depends on the bins size
(Appendix\,\ref{sec:Reference-values-for}), and doe not have a predefined
reference value. As observed for the ZVE defined by Pernot\citep{Pernot2023a_arXiv},
one might expect the ZMSE to be more reliable than the ENCE for heavy-tailed
uncertainty distributions, but it shares its sensitivity to heavy-tailed
error distributions\citep{Pernot2024_arXiv}.

\subsection{Correlation}

The correlation between absolute errors and uncertainties 
\begin{equation}
CC=\mathrm{cor}(|E|,u_{E})
\end{equation}
is used to assess the strength of the link between these variables\citep{Tynes2021}.
One expects large absolute errors to be associated with large uncertainties
and small uncertainties to be associated with small absolute errors.
However, the link is not symmetric, as small absolute errors might
be associated with large uncertainties. To account for a possible
non linear relation and to reduce the sensitivity to outliers, \emph{Spearman's}
\emph{rank correlation coefficient} is recommended to estimate CC. 

A positive value of CC is desirable, but, considering the probabilistic
link between errors and uncertainties, its reference value cannot
be 1.\citep{Pernot2022b} In absence of a specified target value,
the CC alone should not be used in comparative studies, nor for validation. 

\subsection{Validation\label{subsec:Validation-approaches-for}}

The validation protocol has been presented in Ref\citep{Pernot2024_arXiv}.
The main points are summarized here. For a given dataset $(E,u_{E})$
and statistic $\vartheta$, one estimates the statistic over the dataset
$\vartheta_{est}$ and a 95\,\% confidence interval $I_{BS}=\left[I_{BS}^{-},I_{BS}^{+}\right]$
by bootstrapping using the Bias Corrected Accelerated (BC$_{a}$)
method\citep{DiCiccio1996}. One can then test that the target value
for the statistic, $\vartheta_{ref}$, lies within $I_{BS}$, i.e.
\begin{equation}
\vartheta_{ref}\in\left[I_{BS}^{-},I_{BS}^{+}\right]\label{eq:int-valid}
\end{equation}
For a non-binary agreement measure, a standardized score $\zeta$
is defined as
\begin{equation}
\zeta(\vartheta_{est},\vartheta_{ref},I_{BS})=\begin{cases}
\frac{\vartheta_{est}-\vartheta_{ref}}{I_{BS}^{+}-\vartheta_{est}} & if\,(\vartheta_{est}-\vartheta_{ref})\le0\\
\frac{\vartheta_{est}-\vartheta_{ref}}{\vartheta_{est}-I_{BS}^{-}} & if\,(\vartheta_{est}-\vartheta_{ref})>0
\end{cases}\label{eq:zeta-def}
\end{equation}
that can be tested by
\begin{equation}
|\zeta(\vartheta_{est},\vartheta_{ref},I_{BS})|\le1\label{eq:zeta-valid}
\end{equation}
which is equivalent to the interval test (Eq.\,\ref{eq:int-valid}).
In addition, $\zeta$ provides valuable information about the sign
and amplitude of the agreement between the statistic and its reference
value.

The validation procedure depends then on the availability of $\vartheta_{ref}$,
as shown next.

\subsubsection{Predefined $\vartheta_{ref}$\label{subsec:Predefined}}

If $\vartheta_{ref}$ is known, one can directly use Eq.~\ref{eq:zeta-def}
to estimate 
\begin{equation}
\zeta_{BS}=\zeta(\vartheta_{est},\vartheta_{ref},I_{BS})\label{eq:benchmark}
\end{equation}
which will be considered below as the benchmark method against which
the alternative simulation-based methods will be evaluated.

\subsubsection{Estimation of $\vartheta_{ref}$ by simulation\label{par:Estimation-of-theta-ref}}

For those statistics without a predefined reference value, one has
to make an hypothesis on $D$, in order to generate $\vartheta_{ref}$
from ideally calibrated datasets:
\begin{enumerate}
\item Choose a unit-variance generative distribution $D(0,1)$.
\item Draw a set of pseudo-errors from the actual uncertainties by applying
Eq.\,\ref{eq:probmod} and estimate the corresponding statistic $\tilde{\vartheta}_{D}$,
where the subscript $D$ denotes the choice of generative distribution.
\item Repeat step 2 $N_{MC}$ times (Monte Carlo sampling).
\item Take the mean value $\tilde{\vartheta}_{D,ref}=<\tilde{\vartheta}_{D}>$
and estimate the \emph{standard error} on $\tilde{\vartheta}_{D,ref}$:
$u(\tilde{\vartheta}_{D,ref})=\mathrm{sd}(\tilde{\vartheta}_{D})/\sqrt{N_{MC}}$,
where $\mathrm{sd}(x)$ is the standard deviation of variable $x$.
\end{enumerate}
For $N_{MC}$ on the order of $10^{4}$, one should have $2u(\tilde{\vartheta}_{D,ref})\ll U_{BS}$,
where $U_{BS}$ is the half range of $I_{BS}$, and Eq.\,\ref{eq:zeta-def}
can be applied without accounting for the uncertainty on $\tilde{\vartheta}_{D,ref}$.
One can then estimate a $\zeta$-score
\begin{equation}
\zeta_{SimD}=\zeta(\vartheta_{est},\tilde{\vartheta}_{D,ref},I_{BS})\label{eq:simrefval}
\end{equation}

Note that this approach differs from the one proposed previously in
the literature\citep{Pernot2022c,Rasmussen2023}, where the value
of the statistic was considered as fixed and the reference value as
a random variable with uncertainty $\mathrm{sd}(\tilde{\vartheta}_{D})$.
This scenario is implemented for comparison, with 
\begin{equation}
\zeta_{Sim2D}=\zeta(\vartheta_{est},\tilde{\vartheta}_{D,ref},I_{D})
\end{equation}
using a confidence interval $I_{D}$ estimated from the the quantiles
of the simulated sample of $\tilde{\vartheta}_{D}$.

It is important to acknowledge that validation tests based on the
latter approach present two drawbacks when compared to the bootstrapping-based
procedure: (1) they ignore uncertainty on $\vartheta_{est}$, creating
an asymmetric treatment of the statistics, depending on the existence
or not of a known reference value; (2) the limits of the confidence
interval $I_{D}$ can be very sensitive to the choice of generative
distribution $D$,\citep{Pernot2022c} as will be illustrated in Sect.\,\ref{sec:Applications}.

\subsubsection{Recommended validation workflow}

\begin{figure}[t]
\begin{centering}
\includegraphics[viewport=0bp 0bp 1000bp 1123bp,clip,width=0.8\textwidth]{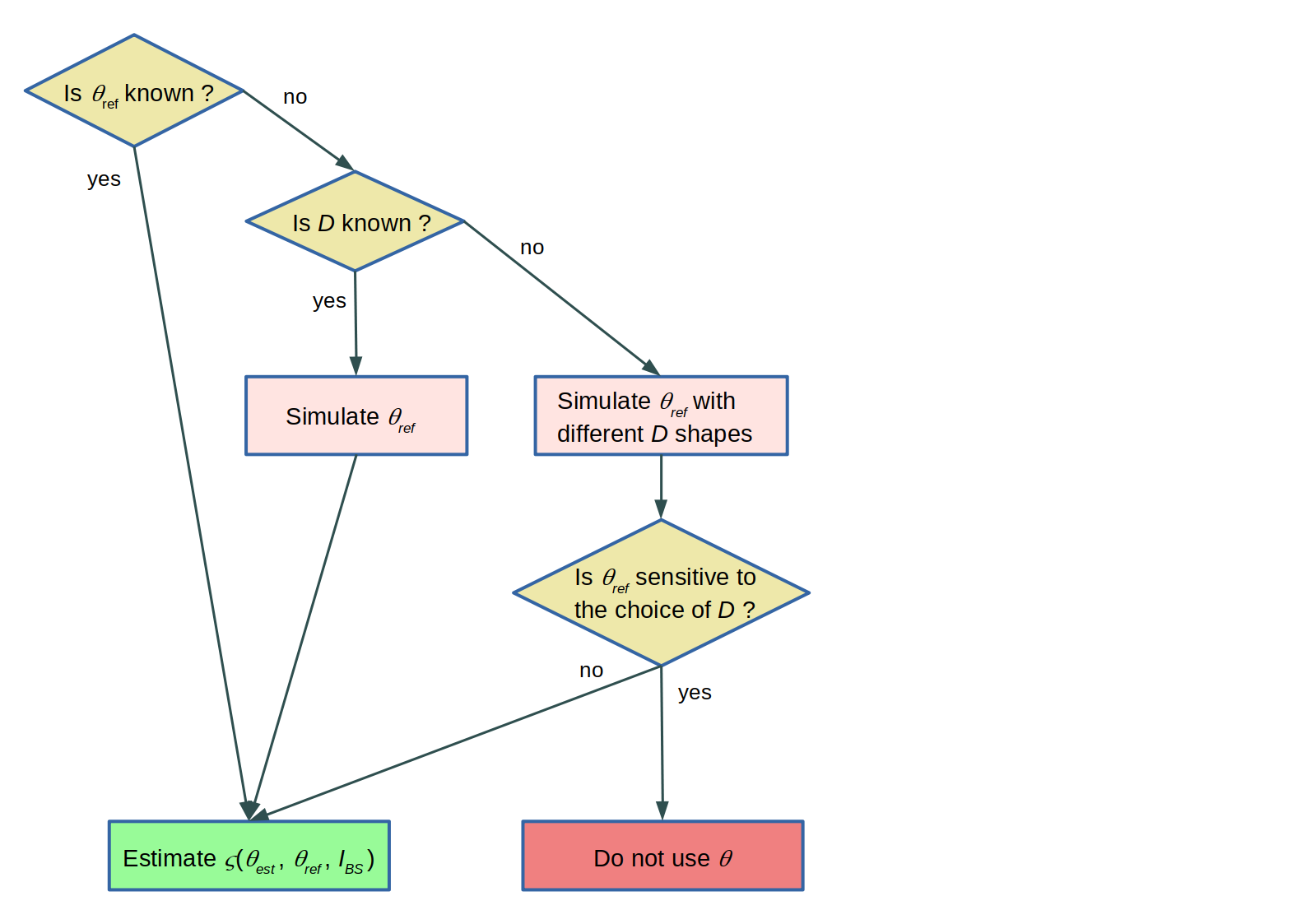}
\par\end{centering}
\caption{\label{fig:Flowchart-for-the}Flowchart for the validation of a statistic
$\vartheta$. $\vartheta_{ref}$ is the reference value used for validation,
$\vartheta_{est}$ is the actual value of the statistic, $I_{BS}$
is the bootstrapped CI for $\vartheta_{est}$ and $D$ is the error
generative distribution.}
\end{figure}

A recommended validation workflow for a given statistic $\vartheta$
is shown in Fig.\,\ref{fig:Flowchart-for-the}. Note that it is essential
that $\vartheta$ should be first confirmed as being fit to the purpose
of the study, notably for datasets with highly skewed and/or heavy-tailed
uncertainty and/or error distributions\citep{Pernot2024_arXiv}. 

For statistics with a known reference value, one can apply directly
the benchmark method (Eq.\,\ref{eq:benchmark}). In the absence of
a predefined reference value for a statistic, one might generate a
simulated one (Sect.\,\ref{par:Estimation-of-theta-ref}), but a
crucial point of this method is the choice of a generative distribution
$D$ linking the errors to the uncertainties (Eq.\,\ref{eq:probmod}). 

If $D$ is well constrained, one may proceed to the estimation of
the simulated reference value $\tilde{\vartheta}_{D,ref}$ and use
it for validation (Eq.\,\ref{eq:simrefval}). In absence of constraints
on $D,$ it is essential to estimate the sensitivity of $\tilde{\vartheta}_{D,ref}$
to $D$. For this, at least two alternative shapes for $D$ have to
be considered, for instance $D=N(0,1)$ and $D=t_{s}(\nu)$, where
the unit-variance Student's-\emph{t} distribution is defined as
\begin{equation}
t_{s}(\nu)=t(\nu)/\sqrt{\nu/(\nu-2)}
\end{equation}
One should use a value of $\nu$ small enough to provide a contrast
with the normal distribution, but not too small, as it might generate
problematic error sets with very heavy tails and many outliers. From
numerical experiments in a former study\citep{Pernot2024_arXiv},
a value of $\nu=6$ is found to be a good compromise. If the simulated
reference values for these choices of $D$ differ more than their
statistical uncertainty, the statistic is deemed to be over-sensitive
to $D$ and should not be used for validation. Otherwise, one might
proceed as in the case of a predefined $D$ (Eq.\,\ref{eq:simrefval}).

\section{Applications\label{sec:Applications}}

The validation approach presented above is applied to nine datasets
extracted from the ML-UQ literature, with a focus on the shape of
\emph{z}-scores distributions and its consequences on the validation
process. This is followed by a thorough study of the sensitivity of
the validation procedure to various parameters, such as the uncertainty
distribution and the shape of the generative distribution.

\subsection{The datasets}

Nine test sets with \emph{a priori} calibrated uncertainties have
been extracted from the recent ML-UQ literature for the prediction
of physico-chemical properties by a diverse panel of ML and UQ methods.
These datasets were tested by Pernot\citep{Pernot2024_arXiv} for
average calibration by the RCE and ZMS statistics. The characterization
of the squared uncertainty and squared error distributions by robust
skewness and kurtosis statistics ($\beta_{GM}$ and $\kappa_{CS}$)\citep{Crow1967,Groeneveld1984,Bonato2011,Pernot2021}
was used to screen heavy-tailed distributions likely to cause reliability
problems in the statistical validation of RCE and ZMS. It was found
that $\beta_{GM}$ and $\kappa_{CS}$ were mostly redundant for these
datasets, and only $\beta_{GM}$ is reported in Table\,\ref{tab:data}.
$\beta_{GM}$ is robust to outliers, varies between -1 and 1 and is
null for symmetric distributions. For $u_{E}^{2}$ distributions,
an upper safety limit of $0.6$ for $\beta_{GM}$ was established
based on an Inverse Gamma distribution model, while for $E^{2}$ and
$Z^{2}$ this limit is 0.8, based on a Fisher distribution model\citep{Pernot2024_arXiv}.
Values exceeding these limits are noted in bold face in Table\,\ref{tab:data}.
\begin{table}[t]
\noindent \begin{centering}
\begin{tabular}{clrrccc}
\hline 
Set \#  & Name & Size ($M$) &  & {\small{}$\beta_{GM}(u_{E}^{2})$} & {\small{}$\beta_{GM}(E^{2})$} & {\small{}$\beta_{GM}(Z^{2})$}\tabularnewline
\cline{1-3} \cline{5-7} 
1  & Diffusion\_RF\citep{Palmer2022} & 2040  &  & {\small{}0.40} & \textbf{\small{}0.82}{\small{} } & {\small{}0.73}\tabularnewline
2 & Perovskite\_RF\citep{Palmer2022} & 3834  &  & \textbf{\small{}0.72} & \textbf{\small{}0.94}{\small{} } & \textbf{\small{}0.83}\tabularnewline
3 & Diffusion\_LR\citep{Palmer2022}  & 2040  &  & \textbf{\small{}0.66} & {\small{}0.74 } & {\small{}0.69}\tabularnewline
4 & Perovskite\_LR\citep{Palmer2022}  & 3836  &  & \textbf{\small{}0.74} & \textbf{\small{}0.82}{\small{} } & {\small{}0.69}\tabularnewline
5 & Diffusion\_GPR\_Bayesian\citep{Palmer2022}  & 2040  &  & {\small{}0.19} & {\small{}0.78 } & {\small{}0.79}\tabularnewline
6 & Perovskite\_GPR\_Bayesian\citep{Palmer2022}  & 3818  &  & {\small{}0.50} & \textbf{\small{}0.96}{\small{} } & \textbf{\small{}0.95}\tabularnewline
7 & QM9\_E\citep{Busk2022} & 13885  &  & \textbf{\small{}0.93} & \textbf{\small{}0.98}{\small{} } & {\small{}0.78}\tabularnewline
8  & logP\_10k\_a\_LS-GCN\citep{Rasmussen2023}  & 5000  &  & {\small{}0.30} & {\small{}0.79 } & {\small{}0.78}\tabularnewline
9  & logP\_150k\_LS-GCN\citep{Rasmussen2023}  & 5000  &  & {\small{}0.30} & {\small{}0.77 } & {\small{}0.75}\tabularnewline
\hline 
\end{tabular}
\par\end{centering}
\caption{\label{tab:data}The nine datasets used in this study: number, name,
size and reference. }
\end{table}
\textcolor{red}{{} }

In this previous study\citep{Pernot2024_arXiv}, the analysis of $u_{E}^{2}$
and $E^{2}$ distributions enabled also to show that applying the
generative model (Eq.\,\ref{eq:probmod}) led to reject the normality
of $D$ for all the studied datasets. Typically, the errors are much
more dispersed than what would be expected from a normal generative
distribution. As this is a central point for the estimation of simulated
reference values, the distribution of $Z$, which should reflect the
empirical shape of $D$ for calibrated datasets is analyzed next. 

\subsubsection{Distributions of z-scores\label{sec:Errors-and-z-scores}}

\begin{figure}[t]
\noindent \begin{centering}
\includegraphics[width=0.9\textwidth]{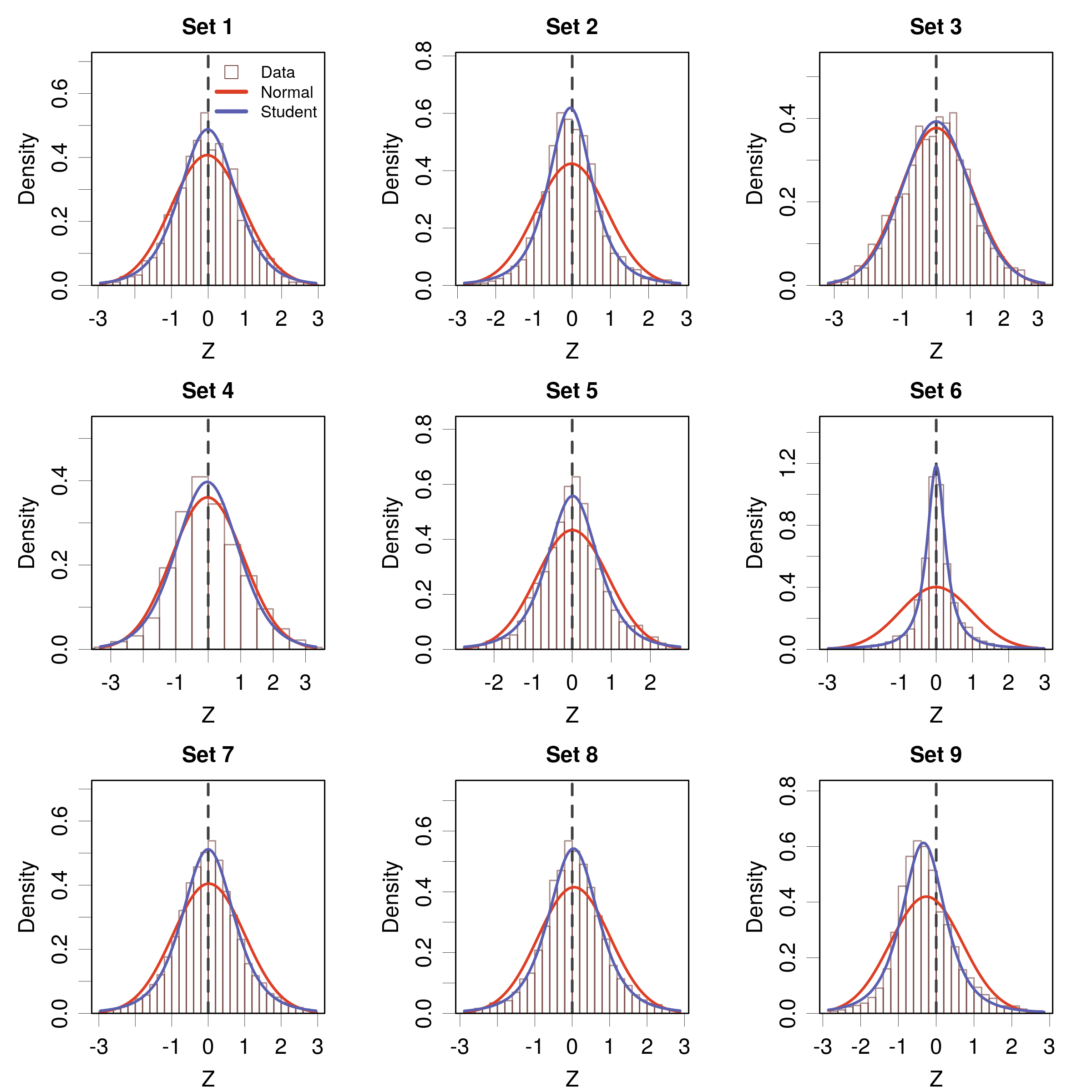}
\par\end{centering}
\caption{\label{fig:Z-dist}Z-scores distributions (histograms) with normal
(red line) and scaled and shifted Student's-\emph{t} (blue line) fits.
For legibility, the histograms have been truncated to $\pm3$ standard
deviations, hiding a few outlying values.}
\end{figure}

The fit of \emph{z}-scores distributions by a scaled-and-shifted Student's-\emph{t}
distribution is done by maximum likelihood estimation \citep{Delignette2015}.\textcolor{orange}{{}
}Table\,\ref{tab:data-summary} reports the mean values, standard
deviations, relative bias and the number of degrees of freedom $\nu_{Z}$
. The smaller $\nu_{Z}$, the farther one is from a normal distribution.
Note that this fit accounts for a possible bias in the dataset, which
is non-negligible for Sets 8 and 9 (above 5\,\% of the standard deviation).
The values of $\nu_{Z}$ reported here are globally consistent with
the ones obtained previously\citep{Pernot2024_arXiv} for the fit
of the distributions of $Z^{2}$ by a Fisher-Snedecor $F(1,\nu_{Z^{2}})$
distribution, except for Set 9 for which the bias was not taken into
account. Fig\,\ref{fig:Z-dist} shows the comparison of the best
fits of \emph{z}-scores by normal and scaled-and-shifted Student's-\emph{t}
distributions.

\noindent 
\begin{table}[t]
\noindent \begin{centering}
\begin{tabular}{ccr@{\extracolsep{0pt}.}lr@{\extracolsep{0pt}.}lr@{\extracolsep{0pt}.}lr@{\extracolsep{0pt}.}lr@{\extracolsep{0pt}.}lr@{\extracolsep{0pt}.}l}
\hline 
Set \#  &  & \multicolumn{2}{c}{$\mu_{Z}$} & \multicolumn{2}{c}{$\sigma_{Z}$} & \multicolumn{2}{c}{$b_{Z}$\,(\%)} & \multicolumn{2}{c}{$\nu_{Z}$} & \multicolumn{2}{c}{} & \multicolumn{2}{c}{$\nu_{Z^{2}}$}\tabularnewline
\cline{1-1} \cline{3-10} \cline{13-14} 
1  &  & -0&027(22)  & 0&980(30)  & \multicolumn{2}{c}{3 } & 6&0 & \multicolumn{2}{c}{} & 7&9 \tabularnewline
2  &  & -0&018(15)  & 0&940(26)  & \multicolumn{2}{c}{2 } & 3&3 & \multicolumn{2}{c}{} & 4&9 \tabularnewline
3  &  & 0&002(23)  & 1&058(18)  & \multicolumn{2}{c}{0 } & 20&1 & \multicolumn{2}{c}{} & 15&1 \tabularnewline
4  &  & -0&021(18)  & 1&107(16)  & \multicolumn{2}{c}{2 } & 9&1 & \multicolumn{2}{c}{} & 8&2 \tabularnewline
5  &  & 0&006(20)  & 0&920(21)  & \multicolumn{2}{c}{1 } & 3&9 & \multicolumn{2}{c}{} & 2&7 \tabularnewline
6  &  & -0&005(16)  & 0&992(37)  & \multicolumn{2}{c}{1 } & 1&4 & \multicolumn{2}{c}{} & 0&9 \tabularnewline
7  &  & 0&0174(84)  & 0&9858(99)  & \multicolumn{2}{c}{2 } & 4&4 & \multicolumn{2}{c}{} & 4&0 \tabularnewline
8  &  & 0&050(14)  & 0&961(16)  & \multicolumn{2}{c}{5 } & 3&9 & \multicolumn{2}{c}{} & 3&7 \tabularnewline
9  &  & -0&260(13)  & 0&951(23)  & \multicolumn{2}{c}{27 } & 3&1 & \multicolumn{2}{c}{} & 20&2 \tabularnewline
\hline 
\end{tabular}
\par\end{centering}
\noindent \begin{centering}
\par\end{centering}
\caption{\label{tab:data-summary}Summary statistics for the \emph{z}-scores
($Z$): $\mu_{Z}$ is the mean value, $\sigma_{Z}$ the standard deviation,
$b_{Z}=100(\mu_{Z}-0)/\sigma_{Z}$ measures relative bias and $\nu_{Z}$
is the number of degrees of freedom resulting from the fit of the
data by a scaled and shifted Student's-\emph{t} distribution. The
last column reports the shape parameter for the fit of $Z^{2}$ distributions
by a $F(1,\nu_{Z^{2}})$ distribution\citep{Pernot2024_arXiv}.}
\end{table}

The only sets for which one gets close to normality are Set 3, with
$\nu_{Z}=20.1$, and to a lesser degree Set 4, with $\nu_{Z}=9.1$.
Overall, one has rather small $\nu_{Z}$ values, rejecting unambiguously
the normality of \emph{z}-scores for 7 or 8 out of nine datasets.
As a side effect, let us note that the normality of \emph{z}-scores
should not be used as a calibration criterion, unless $D$ is known
to be normal.

\subsection{Analysis of the generative model}

\begin{figure}[t]
\noindent \begin{centering}
\includegraphics[width=0.9\textwidth]{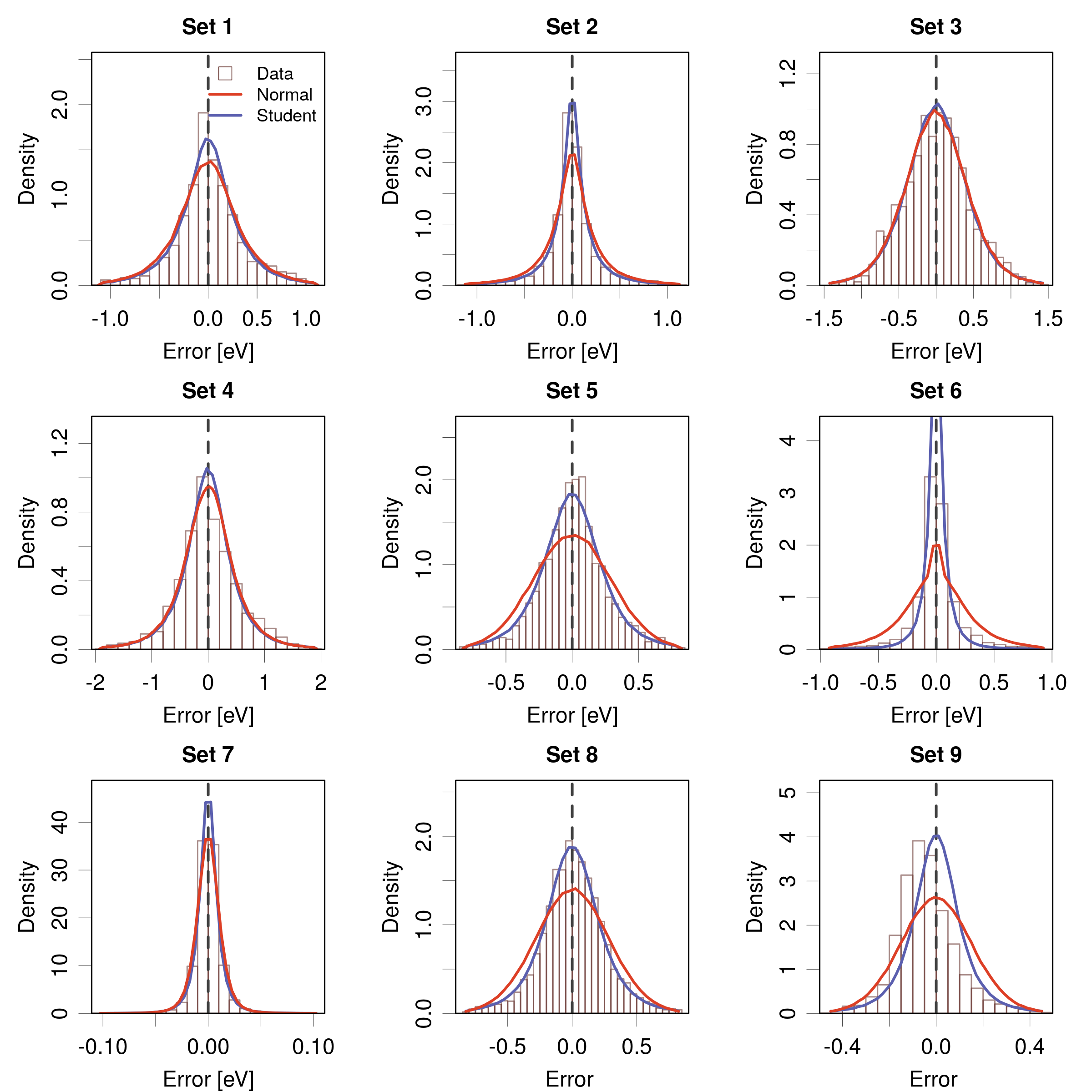}
\par\end{centering}
\caption{\label{fig:Esim-dist}Recovery of the error distributions with the
generative model, using a normal distribution (red line) or a unit-variance
Student's distribution $t_{s}(\nu_{Z})$ ($\nu_{Z}$ from Table\,\ref{tab:data-summary};
blue line). For Set 6, the Student's-\emph{t} model cannot be generated
because of the infinite variance for $\nu_{Z}<2$ and it was replaced
by $\nu_{Z}=2.1$.}
\end{figure}

Considering the generative model, the distribution of \emph{z}-scores
\emph{for calibrated datasets} should be identical to $D$. To check
how close the distribution of \emph{z}-scores is to the errors generative
distribution, simulated error distributions have been generated using
the actual uncertainties and two generative distributions: a standard
Normal distribution $N(0,1)$ and a unit-variance Student's-\emph{t}
distribution $t_{s}(\nu_{Z})$ with the degrees of freedom reported
in Table\,\ref{tab:data-summary}. The results are shown in Fig.\,\ref{fig:Esim-dist}.

The $t_{s}$-based model offers a better fit to the errors histogram
for all Sets except 3 and 4, for which it is on par with the normal
model. Note that for Set 6 the value of $\nu_{Z}$ had to be increased
to 2.1 to avoid the infinite variance due to $\nu_{Z}\le2$. 

It is thus clear that a normal generative distribution $D=N(0,1)$
is not likely to provide simulated error distributions with properties
close to the actual errors, except for a few datasets. 

\subsection{Sensitivity analysis}

In the previous section, we have seen that if there is a best choice
for the generative distribution $D$, it is rarely the normal distribution.
In absence of strong constraints to guide this choice, it is important
to assess the sensitivity to $D$ of the simulated reference values
for the candidate calibration statistics, and more globally the sensitivity
of the validation $\zeta$-scores. 

\subsubsection{Sensitivity of $\tilde{\vartheta}_{D,ref}$ to the uncertainty distribution}

\begin{figure}[t]
\noindent \begin{centering}
\includegraphics[width=0.75\textwidth]{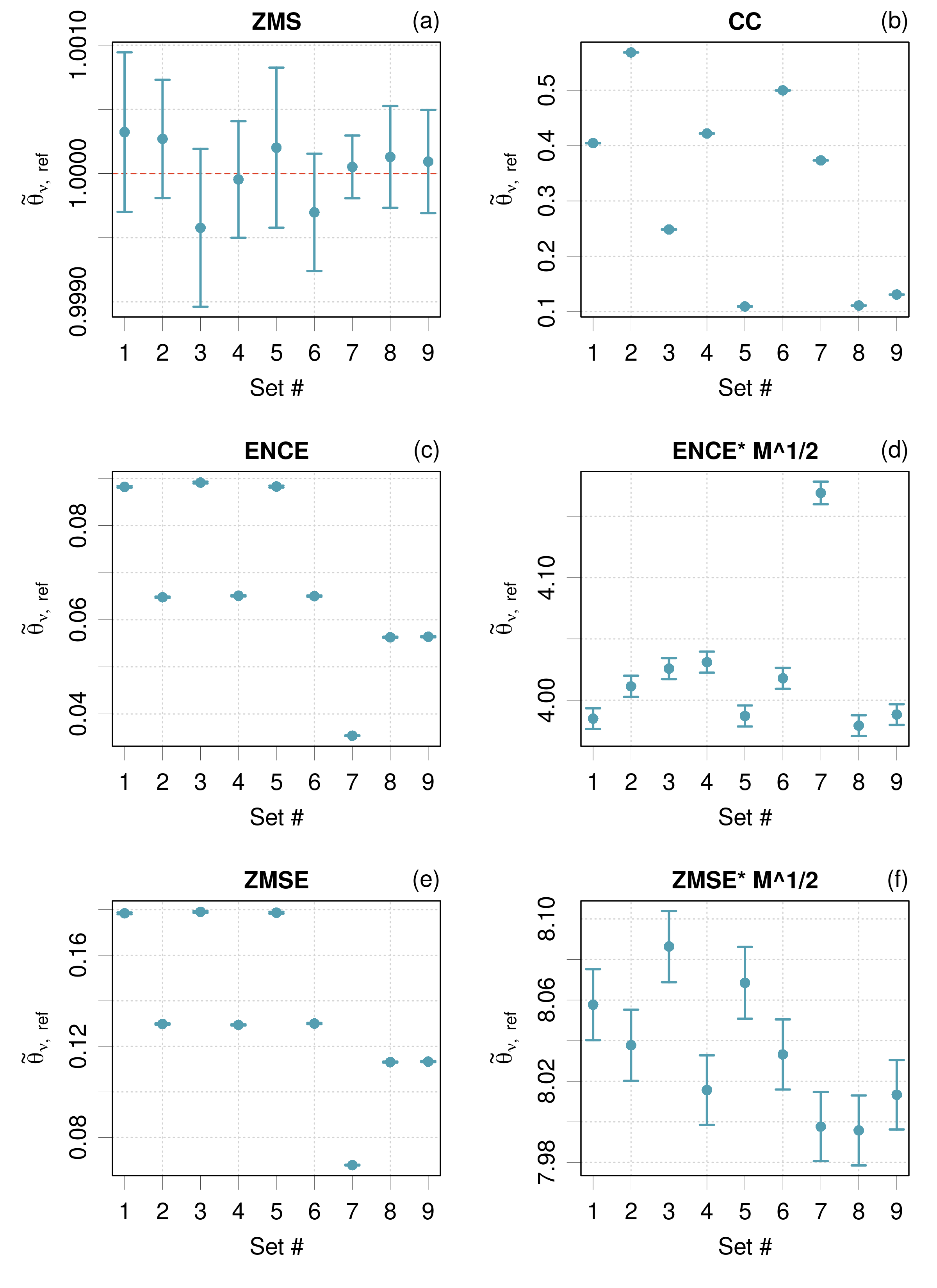}
\par\end{centering}
\caption{\label{fig:sensitivity-1-1}Sensitivity of $\tilde{\vartheta}_{D,ref}$
to the dataset for the ZMS (a), CC (b), ENCE (c, d) and ZMSE (e, f)
statistics. The generative distribution is standard normal, $D=N(0,1)$.}
\end{figure}

A first point is the appreciation of the dependence of the simulated
reference values on the uncertainty distribution. For this, a normal
generative distribution is chosen {[}$D=N(0,1)${]}, and $\tilde{\vartheta}_{D,ref}$
and $u(\tilde{\vartheta}_{D,ref})$ are estimated for each statistic
and the nine example datasets. The sampling size is $N_{MC}=10^{4}$
and $u_{E}$ is sorted and parted into 50 equal-size bins to estimate
the ENCE and ZMSE statistics. The results are presented in Fig.\,\ref{fig:sensitivity-1-1}.

For the ZMS {[}Fig.\,\ref{fig:sensitivity-1-1}(a){]}, the simulation
recovers trivially the predefined reference value, as the simulated
z-scores are directly samples of the $D$ distribution. For CC {[}Fig.\,\ref{fig:sensitivity-1-1}(b){]},
$\tilde{\vartheta}_{D,ref}$ is seen to depend strongly on the dataset,
confirming the absence of a common reference value for these datasets.
The ENCE {[}Fig.\,\ref{fig:sensitivity-1-1}(c){]} and ZMSE {[}Fig.\,\ref{fig:sensitivity-1-1}(e){]}
follow a parallel pattern, which is mostly due to the sensitivity
of these variables to the dataset size, $M$ (Appendix\,\ref{sec:Reference-values-for}).
The correction of this trend by multiplying the statistics by $M^{1/2}$
{[}Fig.\,\ref{fig:sensitivity-1-1}(d,f){]} shows that the common
pattern disappears and that each uncertainty set leads to its own
reference value, even if a few CIs overlap. The ENCE has an outstanding
value of $\tilde{\vartheta}_{D,ref}$ for Set 7, which is not the
case for the ZMSE, and is certainly a symptom of the stratification
of the uncertainties resulting from an isotonic regression post-hoc
calibration\citep{Busk2023,Pernot2023b_arXiv,Pernot2023d}.

Globally, apart from the ZMS, on needs therefore to estimate a simulated
reference value for each set and statistic. 

\subsubsection{Sensitivity of $\tilde{\vartheta}_{D,ref}$ to $D$}

\begin{figure}[t]
\noindent \begin{centering}
\includegraphics[width=0.75\textwidth]{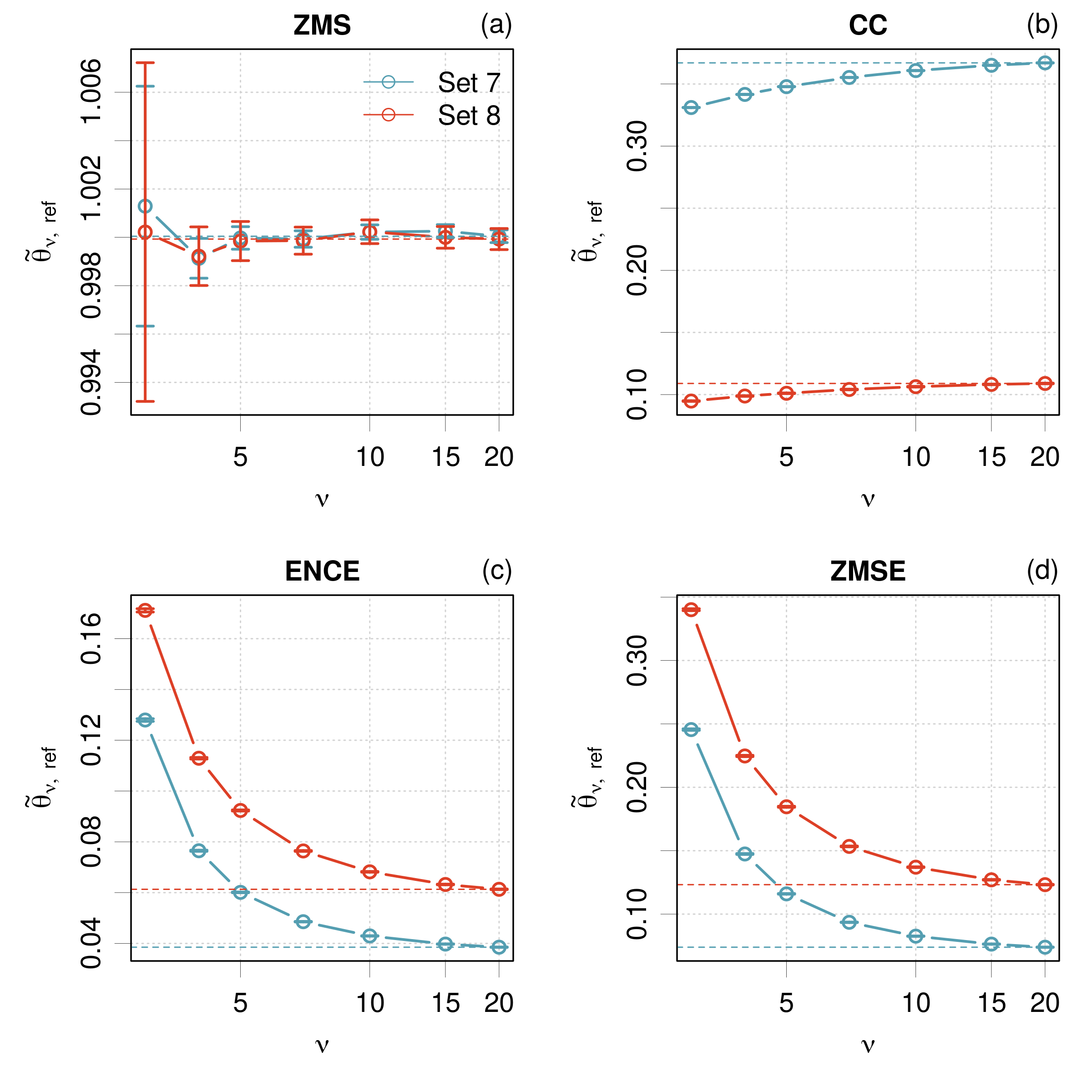}
\par\end{centering}
\caption{\label{fig:sensitivity-1}Sensitivity of the simulated reference values
to the generative distribution. Calibrated datasets are generated
from the actual uncertainties of Sets 7 and 8, using a generative
distribution $D=t_{s}(\nu)$ with $\nu$ degrees of freedom. }
\end{figure}

The simulated reference values are generated for Sets 7 and 8 (as
representative of the largest sets with markedly different uncertainty
distributions) from their actual uncertainties by using a unit-variance
Student's-\emph{t} generative distribution $D=t_{s}(\nu)$ for a range
of degrees of freedom between 3 and 20, covering the range of values
observed for $\nu_{Z}$ (Table\,\ref{tab:data-summary}). The sampling
size is $N_{MC}=10^{4}$ and $u_{E}$ is sorted and parted into 50
equal-size bins to estimate the ENCE, ZMSE. The results are presented
in Fig.\,\ref{fig:sensitivity-1}. 

For ZMS {[}Fig.\,\ref{fig:sensitivity-1}(a){]}, one expects to observe
the conformity of the simulated reference value with the known reference
value (i.e. $\tilde{\vartheta}_{\nu,ref}\simeq1$), independently
of the chosen generative distribution. This is the case, albeit with
a notable increase of Monte Carlo sampling uncertainty as $v$ decreases.
For CC {[}Fig.\,\ref{fig:sensitivity-1}(b){]}, the sensitivity to
$\nu$ depends on the value of the statistic, and it is less marked
for low CC values (Set 8) than for larger ones (Set 7). For the ENCE
{[}Fig.\,\ref{fig:sensitivity-1}(c){]} and ZMSE {[}Fig.\,\ref{fig:sensitivity-1}(d){]}
statistics, there is no ambiguity, as the simulated reference values
depend strongly on $\nu$, with a similar behavior for both statistics
and datasets.

This sensitivity analysis is based on a worst case scenario where
the lower values of $\nu$ generate amounts of outliers that can disturb
the reliability of calibration statistics. It suggests that using
too extreme generative distributions $D$ for the sensitivity analysis
of those statistics without reference values might be counterproductive.
For instance, a Student's-\emph{t} distribution with defined skewness
and kurtosis requires $\nu\ge5$, and in the following one will use
$t_{s}(\nu=6)$ as an alternative to the normal distribution. 

Ignoring small values of $\nu$, one still sees that the CC, ENCE
and ZMSE statistics can hardly be useful in a validation context when
the generative distribution $D$ is unknown.

\subsubsection{Sensitivity of $\zeta$-scores to $D$}

We now consider how the sensitivity of $\tilde{\vartheta}_{D,ref}$
to $D$ propagates to the $\zeta$-scores. For the simulated reference
values, two options for $D$ are considered: a standard normal ($\zeta_{SimN}$,
$\zeta_{Sim2N}$) and unit-variance Student's-\emph{t} distribution
with 6 degrees of freedom $t_{s}(\nu=6)$ ($\zeta_{SimT}$, $\zeta_{Sim2T}$).
This value has been chosen to avoid the perturbation of the statistics
by extreme values as observed in the previous section.

\begin{figure}[t]
\noindent \begin{centering}
\includegraphics[width=0.87\textwidth]{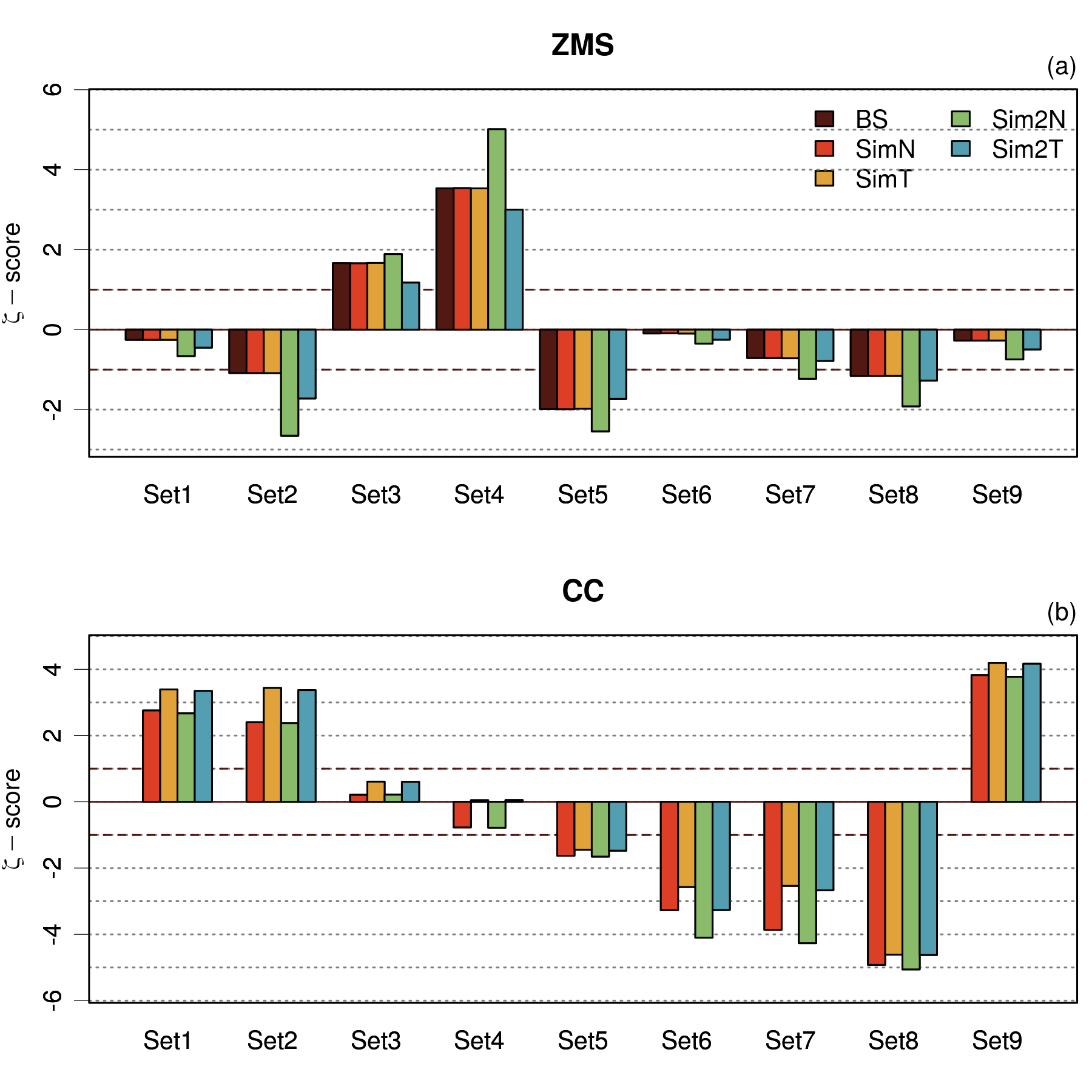}
\par\end{centering}
\caption{\label{fig:z-Scores}Validation of the ZMS and CC statistics with
$\zeta$-scores obtained by bootstrapping (BS) and by simulation with
two hypotheses on the errors distribution $D$ : Normal (SimN) and
$t_{s}(\nu=6)$ (SimT), compared to the Monte Carlo approach of Rasmussen
\emph{et al.}\citep{Rasmussen2023} (Sim2N and Sim2T). In absence
of a predefined reference value, the BS approach is not available
for CC. }
\end{figure}

The results for the ZMS, CC, ENCE and ZMSE statistics and the pertinent
simulation scenarios are reported in Figs.\,\ref{fig:z-Scores}-\ref{fig:z-Scores-1}.
The intermediate variables for the calculation of these $\zeta$-scores
are reported in Appendix\,\ref{sec:Results-tables-for}. 

Let us first consider the results for the ZMS statistic {[}Figs.\,\ref{fig:z-Scores}(a){]}.
There is a good agreement between the three bootstrapped CI-based
estimations (BS, SimN, SimT). The absence of notable difference between
the SimN and SimT results confirms the low sensitivity of this method
to the choice of $D$. It is also clear that the Sim2N and Sim2T scores
can differ notably from the other scores and between them, confirming
a high sensitivity of the Monte Carlo CI $I_{D}$ to $D$. Sets 1-6
have been analyzed in an earlier study\citep{Pernot2023d}, by using
the $\mathrm{Var}(Z)$ score and an approximate analytical estimation
of confidence intervals\citep{Cho2005}. The validation results differ
at the margin, as the approximate method used previously seems to
provide slightly wider CIs than the bootstrapped ones used here, and
is therefore less stringent.

For the CC statistic {[}Figs.\,\ref{fig:z-Scores}(b){]}, it appears
that the choice of $D$ has always a sizable effect, even for the
Sim protocol. In consequence, and unless one trusts more one of the
two options for $D$, the CC statistic cannot be reliably validated
for the studied datasets. Note that, by chance and despite the amplitude
of the $D$-induced discrepancies, all the CC validation statistics
for a given dataset agree on the binary validation diagnostic. One
might be then tempted to conclude that only Sets 3 and 4 have CCs
compatible with their simulated reference values. Note that in this
case, the Sim and Sim2 $\zeta$-scores for a given choice of $D$
are very similar. 
\begin{figure}[t]
\noindent \begin{centering}
\includegraphics[width=0.87\textwidth]{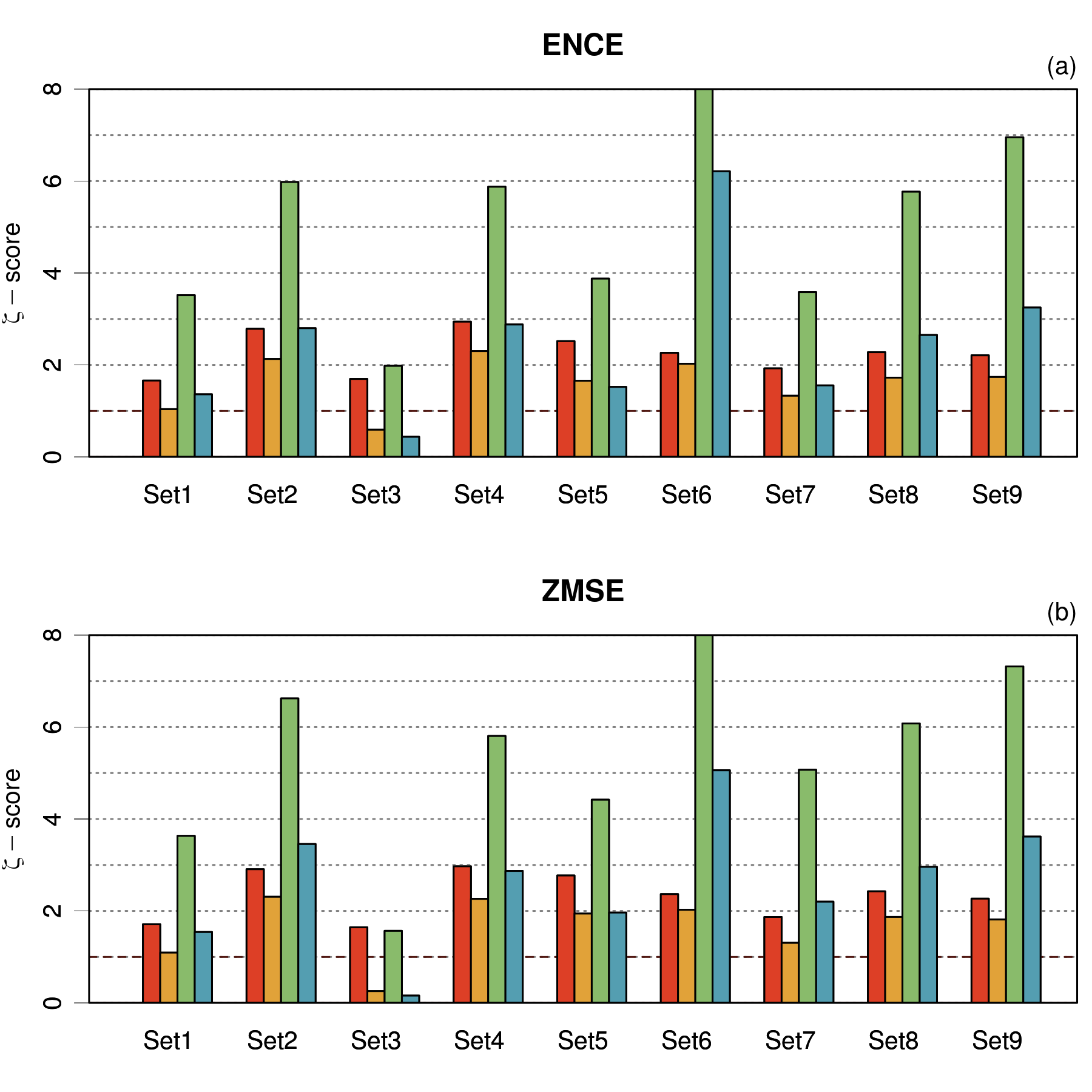}
\par\end{centering}
\caption{\label{fig:z-Scores-1}Same as Fig.\,\ref{fig:z-Scores} for the
ENCE and ZMSE consistency scores. In absence of a predefined reference
value, the BS approach is not available for these statistics. }
\end{figure}

The $\zeta$-scores for the ENCE and ZMSE statistics have been estimated
for $N=20$ bins and present very similar features (Fig.\ref{fig:z-Scores-1}):
a notable sensitivity to $D$ and a large difference between the Sim
and Sim2 protocols (except for Set 3). The latter observation confirms
that one should not use the Monte Carlo CIs to estimate the $\zeta$-scores.
Focusing on the SimN and SimT results, it appears that they globally
agree to reject consistency, except for Set 3 where consistency would
be validated by SimT, as observed in Appendix\,\ref{sec:Validation-of-ZMSE}.
However, we have shown above that for Sets 3 and 4 the SimN scenario
is the most plausible.

It is somewhat striking that the validation scores of ENCE and ZMSE
are so similar, despite notable differences in the reliability of
the RCE and ZMS statistics exposed previously\citep{Pernot2024_arXiv}.
In fact, it was shown that both statistics shared a similar sensitivity
to the tailedness of error distributions, unlike their sensitivity
to the tailedness of uncertainty distributions, where only the RCE
could be notably biased. Binning along $u_{E}$ is expected to erase
the tailedness problem of (binned) uncertainty distributions, placing
both ENCE and ZMSE on an equal footing. This would not be the case
for another binning variable.

\section{Discussion\label{sec:Discussion}}

This study focuses on the validation of correlation and calibration
statistics by standardized $\zeta$-scores, for which one needs reference
values and confidence interval estimates. Reference values are predefined
for the ZMS statistic, but not for CC, ENCE and ZMSE. In the latter
cases, it is possible to generate reference values by a Monte Carlo
approach, which implies to generate simulated error sets from the
actual uncertainties, using a generative distribution $D$. A reference
value and its standard error can be obtained by repeating this random
sampling. The choice of an interval-based validation method results
from the often marked non-normality of the error and uncertainty datasets,
which prevents the use of more standard $z$-scores or $t$-scores.
CIs are therefore estimated by bootstrapping for all statistics, by
taking care that the bootstrap distributions can be non-normal and
asymmetric. 

In absence of a predefined reference, Pernot\citep{Pernot2022c} and
Rasmussen \emph{et al.}\citep{Rasmussen2023} proposed to use the
standard deviation or CI of the simulated reference value as a metric
to compare the estimated statistic to its reference value. This study
shows unambiguously that several disadvantages come with this approach:
(i) it introduces an asymmetry of treatment between the statistics
with known and unknown reference values, using bootstrapping-based
uncertainties for the former and simulation-based uncertainties for
the latter\citep{Rasmussen2023}; (ii) the standard deviation (or
CI) of the simulated reference values is strongly dependent on the
chosen generative distribution $D$; and (iii) in such cases, validation
becomes dependent on the often \emph{ad hoc} choice of $D$. 

To circumvent these issues, it is suggested here to \emph{use the
bootstrapping-based intervals} independently of the existence of a
reference value, and to \emph{characterize the simulated reference
by its} \emph{standard error} (standard deviation of the mean), which
is much less sensitive to the choice of $D$. Moreover, with a moderate
effort on the Monte Carlo sampling, it is possible to ensure that
this standard error becomes negligible before the bootstrapping-based
uncertainty, which solves the problem of the dependence of the $\zeta$-scores
on the simulated reference uncertainty. The impact of this approach
has been shown by comparing SimN to SimT and Sim2N to Sim2T (Figs.\ref{fig:z-Scores}-\ref{fig:z-Scores-1}):
in all cases the proposed approach reduces the sensitivity to $D$,
except for CC where it is about identical. 

Another and less manageable source of sensitivity of the $\zeta$-scores
to $D$ comes with the simulated reference value itself. For instance,
the mean value of any statistic that involves a sign loss of the simulated
errors (e.g. absolute value, squared value...) is likely to be affected
by the shape/width of the errors distribution. This is particularly
visible for the CC, ENCE and ZMSE statistics, but much less for the
ZMS. For confidence curves, Pernot\citep{Pernot2022c} has shown that
the simulated reference based on the RMS of the errors was insensitive
to $D$, which was not the case for the MAE. \emph{The sensitivity
of a simulated reference value to the generative model has therefore
to be directly assessed before using it for validation}.\textcolor{orange}{{} }

\section{Conclusion\label{sec:Conclusion}}

This study sheds a critical look at the validation of an average calibration
statistics (ZMS), a correlation statistic (CC) and two conditional
calibration statistics (ENCE, ZMSE). A benchmark method was defined
for those statistics with a predefined reference value (ZMS) and compared
to simulation approaches for statistics with no predefined reference
value (CC, ENCE and ZMSE). A validation workflow was proposed to deal
with the main obstacles of the simulation approach (Fig.\,\ref{fig:Flowchart-for-the})
and notably with the sensitivity of simulated reference values to
the generative distribution $D$ used to generate ideal synthetic
errors from actual uncertainties.

Important conclusions about the studied statistics have been obtained
from a series of ML-UQ datasets from the literature.
\begin{itemize}
\item \emph{The generative distribution $D$ is not necessarily normal}.
The error distributions obtained by the generative model with a normal
distribution $D$ (Eq.\,\ref{eq:probmod}) rarely agree with the
actual error distributions. Better fits are often obtained by using
the distribution of scaled errors (\emph{z}-scores) as a proxy. In
some instances, e.g. Bayesian Neural Networks or Gaussian Processes,
the generative distributions is prescribed, but this is not generally
the case. Thus, before using an hypothetical generative distribution
to simulate unknown reference values, one should test its pertinence
for the dataset(s) under scrutiny. 
\item \emph{The use of simulated reference values for the CC, ENCE and ZMSE
statistics is not reliable} \emph{if $D$ is unknown}. It has been
shown that the simulated reference values for these statistics are
highly sensitive to the choice of generative distribution $D$. If
$D$ is unknown, one cannot estimate a reliable reference value for
their validation. For CC, checking that it is positive is the best
that can be done and there is not much sense in comparing CC values
of different datasets (the largest value is not necessarily the better!).
A much more powerful and reliable approach is the plotting of RMSE-based
\emph{confidence curves} with bootstrapped CI and simulated reference
value. Similarly, the simulated reference values for measures such
as the ENCE and ZMSE are very sensitive to the choice of $D$ and
therefore cannot generally be used for validation of conditional calibration.
When used for comparative studies, they should be limited to datasets
of identical sizes and to identical binning schemes. Validation is
possible by alternative approaches, such as the ``extrapolation to
zero bins'' method used in Appendix\,\ref{sec:Validation-of-ZMSE},
or maybe by the estimation of the ``fraction of valid bins'' proposed
by Pernot\citep{Pernot2023d}. However, the reliability of the latter
approach still needs to be demonstrated for a diversity of datasets. 
\end{itemize}
Finally, the only case where the use of simulated reference values
seems well adapted is for confidence curves, as the RMSE-based reference
curve does not depend on $D$. However, there are probably other calibration
statistics for which this concept might be useful, and the proposed
workflow should enable to validate them with confidence.

\section*{Acknowledgments}

\noindent I warmly thank J. Busk for providing the QM9 dataset.

\section*{Author Declarations}

\subsection*{Conflict of Interest}

The author has no conflicts to disclose.

\section*{Code and data availability\label{sec:Code-and-data}}

\noindent The code and data to reproduce the results of this article
are available at \url{https://github.com/ppernot/2024_SimRef/releases/tag/v1.0}
and at Zenodo (\url{https://doi.org/10.5281/zenodo.10730985}). 

\bibliographystyle{unsrturlPP}
\bibliography{NN}

\begin{thebibliography}{10}

\bibitem{Tynes2021}
M.~Tynes, W.~Gao, D.~J. Burrill, E.~R. Batista, D.~Perez, P.~Yang, and
  N.~Lubbers.
\newblock \href{http://dx.doi.org/10.1021/acs.jcim.1c00670}{Pairwise difference
  regression: A machine learning meta-algorithm for improved prediction and
  uncertainty quantification in chemical search}.
\newblock {\em J. Chem. Inf. Model.}, 61:3846--3857, 2021.
\newblock PMID: 34347460.

\bibitem{Pernot2022b}
P.~Pernot.
\newblock \href{http://dx.doi.org/10.1063/5.0109572}{Prediction uncertainty
  validation for computational chemists}.
\newblock {\em J. Chem. Phys.}, 157:144103, 2022.

\bibitem{Levi2022}
D.~Levi, L.~Gispan, N.~Giladi, and E.~Fetaya.
\newblock \href{http://dx.doi.org/10.3390/s22155540}{{Evaluating and
  Calibrating Uncertainty Prediction in Regression Tasks}}.
\newblock {\em Sensors}, 22:5540, 2022.

\bibitem{Pernot2023a_arXiv}
P.~Pernot.
\newblock \href{http://dx.doi.org/10.48550/arXiv.2305.11905}{Properties of the
  {ENCE} and other {MAD}-based calibration metrics}.
\newblock {\em arXiv:2305.11905}, May 2023.

\bibitem{Pernot2022c}
P.~Pernot.
\newblock \href{http://dx.doi.org/10.48550/arXiv.2206.15272}{{Confidence curves
  for UQ validation: probabilistic reference vs. oracle}}.
\newblock {\em arXiv:2206.15272}, June 2022.

\bibitem{Rasmussen2023}
M.~H. Rasmussen, C.~Duan, H.~J. Kulik, and J.~H. Jensen.
\newblock \href{http://dx.doi.org/10.1186/s13321-023-00790-0}{{Uncertain of
  uncertainties? A comparison of uncertainty quantification metrics for
  chemical data sets}}.
\newblock {\em J. Cheminf.}, 15:1--17, December 2023.

\bibitem{Tran2020}
K.~Tran, W.~Neiswanger, J.~Yoon, Q.~Zhang, E.~Xing, and Z.~W. Ulissi.
\newblock \href{http://dx.doi.org/10.1088/2632-2153/ab7e1a}{Methods for
  comparing uncertainty quantifications for material property predictions}.
\newblock {\em Mach. Learn.: Sci. Technol.}, 1:025006, 2020.

\bibitem{GUM}
{BIPM}, {IEC}, {IFCC}, {ILAC}, {ISO}, {IUPAC}, {IUPAP}, and {OIML}.
\newblock
  \href{http://www.bipm.org/utils/common/documents/jcgm/JCGM_100_2008_F.pdf}{Evaluation
  of measurement data - {G}uide to the expression of uncertainty in measurement
  ({GUM})}.
\newblock Technical Report 100:2008, Joint Committee for Guides in Metrology,
  JCGM, 2008.
\newblock URL:
  \url{http://www.bipm.org/utils/common/documents/jcgm/JCGM_100_2008_F.pdf}.

\bibitem{Busk2022}
J.~Busk, P.~B. J{\o}rgensen, A.~Bhowmik, M.~N. Schmidt, O.~Winther, and
  T.~Vegge.
\newblock \href{http://dx.doi.org/10.1088/2632-2153/ac3eb3}{Calibrated
  uncertainty for molecular property prediction using ensembles of message
  passing neural networks}.
\newblock {\em Mach. Learn.: Sci. Technol.}, 3:015012, 2022.

\bibitem{Amini2019}
A.~Amini, W.~Schwarting, A.~Soleimany, and D.~Rus.
\newblock \href{http://dx.doi.org/10.48550/arXiv.1910.02600}{{Deep Evidential
  Regression}}.
\newblock {\em arXiv:1910.02600}, October 2019.

\bibitem{Evans2000}
M.~Evans, N.~Hastings, and B.~Peacock.
\newblock {\em Statistical Distributions}.
\newblock Wiley-Interscience, 3rd edition, 2000.

\bibitem{Pernot2024_arXiv}
P.~Pernot.
\newblock \href{https://arxiv.org/abs/2402.10043}{{Negative impact of
  heavy-tailed uncertainty and error distributions on the reliability of
  calibration statistics for machine learning regression tasks}}.
\newblock {\em arXiv:2402.10043}, February 2024.
\newblock URL: \url{https://arxiv.org/abs/2402.10043}.

\bibitem{Pernot2023d}
P.~Pernot.
\newblock \href{http://dx.doi.org/10.1063/5.0174943}{{Calibration in machine
  learning uncertainty quantification: Beyond consistency to target
  adaptivity}}.
\newblock {\em APL Mach. Learn.}, 1:046121, 2023.

\bibitem{Pernot2022a}
P.~Pernot.
\newblock \href{http://dx.doi.org/10.1063/5.0084302}{The long road to
  calibrated prediction uncertainty in computational chemistry}.
\newblock {\em J. Chem. Phys.}, 156:114109, 2022.

\bibitem{Busk2023}
J.~Busk, M.~N. Schmidt, O.~Winther, T.~Vegge, and P.~B. J{\o}rgensen.
\newblock \href{http://dx.doi.org/10.1039/D3CP02143B}{{Graph neural network
  interatomic potential ensembles with calibrated aleatoric and epistemic
  uncertainty on energy and forces}}.
\newblock {\em Phys. Chem. Chem. Phys.}, 25:25828--25837, 2023.

\bibitem{Zhang2023}
W.~Zhang, Z.~Ma, S.~Das, T.-W. Weng, A.~Megretski, L.~Daniel, and L.~M. Nguyen.
\newblock \href{http://dx.doi.org/10.48550/arXiv.2312.10469}{{One step closer
  to unbiased aleatoric uncertainty estimation}}.
\newblock {\em arXiv:2312.10469}, December 2023.

\bibitem{Gneiting2007a}
T.~Gneiting, F.~Balabdaoui, and A.~E. Raftery.
\newblock
  \href{http://dx.doi.org/https://doi.org/10.1111/j.1467-9868.2007.00587.x}{Probabilistic
  forecasts, calibration and sharpness}.
\newblock {\em J. R. Statist. Soc. B}, 69:243--268, 2007.

\bibitem{Pernot2023c_arXiv}
P.~Pernot.
\newblock \href{http://dx.doi.org/10.48550/arXiv.2310.11978}{{Can bin-wise
  scaling improve consistency and adaptivity of prediction uncertainty for
  machine learning regression ?}}
\newblock {\em arXiv:2310.11978}, October 2023.

\bibitem{DiCiccio1996}
T.~J. DiCiccio and B.~Efron.
\newblock \href{https://www.jstor.org/stable/2246110}{Bootstrap confidence
  intervals}.
\newblock {\em Statist. Sci.}, 11:189--212, 1996.
\newblock URL: \url{https://www.jstor.org/stable/2246110}.

\bibitem{Crow1967}
E.~L. Crow and M.~M. Siddiqui.
\newblock \href{http://dx.doi.org/10.2307/2283968}{Robust estimation of
  location}.
\newblock {\em J. Am. Stat. Assoc.}, 62:353--389, 1967.

\bibitem{Groeneveld1984}
R.~A. Groeneveld and G.~Meeden.
\newblock \href{http://dx.doi.org/10.2307/2987742}{Measuring skewness and
  kurtosis}.
\newblock {\em The Statistician}, 33:391--399, 1984.
\newblock URL: \url{http://www.jstor.org/stable/2987742}.

\bibitem{Bonato2011}
M.~Bonato.
\newblock \href{http://dx.doi.org/10.1016/j.frl.2010.12.001}{Robust estimation
  of skewness and kurtosis in distributions with infinite higher moments}.
\newblock {\em Financ Res Lett}, 8:77--87, 2011.

\bibitem{Pernot2021}
P.~Pernot and A.~Savin.
\newblock \href{http://dx.doi.org/10.1007/s00214-021-02725-0}{Using the {Gini}
  coefficient to characterize the shape of computational chemistry error
  distributions}.
\newblock {\em Theor. Chem. Acc.}, 140:24, 2021.

\bibitem{Palmer2022}
G.~Palmer, S.~Du, A.~Politowicz, J.~P. Emory, X.~Yang, A.~Gautam, G.~Gupta,
  Z.~Li, R.~Jacobs, and D.~Morgan.
\newblock \href{http://dx.doi.org/10.1038/s41524-022-00794-8}{{Calibration
  after bootstrap for accurate uncertainty quantification in regression
  models}}.
\newblock {\em npj Comput. Mater.}, 8:115, 2022.

\bibitem{Delignette2015}
M.~L. Delignette-Muller and C.~Dutang.
\newblock \href{http://dx.doi.org/10.18637/jss.v064.i04}{{fitdistrplus}: An {R}
  package for fitting distributions}.
\newblock {\em J Stat Softw}, 64(4):1--34, 2015.

\bibitem{Pernot2023b_arXiv}
P.~Pernot.
\newblock \href{http://dx.doi.org/10.48550/arXiv.2306.05180}{{Stratification of
  uncertainties recalibrated by isotonic regression and its impact on
  calibration error statistics}}.
\newblock {\em arXiv:2306.05180}, June 2023.

\bibitem{Cho2005}
E.~Cho, M.~J. Cho, and J.~Eltinge.
\newblock \href{http://www.ijpam.eu/contents/2005-21-3/10/10.pdf}{The variance
  of sample variance from a finite population}.
\newblock {\em Int. J. Pure Appl. Math.}, 21:387--394, 2005.
\newblock URL: \url{http://www.ijpam.eu/contents/2005-21-3/10/10.pdf}.

\end{thebibliography}

\clearpage{}

\appendix

\section*{Appendices}

\beginappendix

\section{Results tables for standardized scores\label{sec:Results-tables-for}}

The numerical values necessary to build Fig.\,\ref{fig:z-Scores}
are reported here (Tables\,\ref{tab:ZMS}-\ref{tab:ZMSE}), with
a summary of the necessary equations. Each table contains the values
for one of the ZMS and CC statistics. 
\begin{table}[t]
\noindent \begin{centering}
{\footnotesize{}}%
\begin{turn}{90}
\begin{tabular}{r@{\extracolsep{0pt}.}lr@{\extracolsep{0pt}.}lr@{\extracolsep{0pt}.}lr@{\extracolsep{0pt}.}lr@{\extracolsep{0pt}.}lr@{\extracolsep{0pt}.}lr@{\extracolsep{0pt}.}lr@{\extracolsep{0pt}.}lr@{\extracolsep{0pt}.}lcr@{\extracolsep{0pt}.}lr@{\extracolsep{0pt}.}lcr@{\extracolsep{0pt}.}lr@{\extracolsep{0pt}.}lr@{\extracolsep{0pt}.}lr@{\extracolsep{0pt}.}lr@{\extracolsep{0pt}.}lc}
\hline 
\multicolumn{2}{|c|}{$\vartheta=ZMS$~} & \multicolumn{2}{c}{} & \multicolumn{2}{c}{} & \multicolumn{8}{c}{BS} & \multicolumn{2}{c}{} & \multicolumn{8}{c}{SimN, Sim2N} & \multicolumn{2}{c}{} & \multicolumn{9}{c}{SimT, Sim2T}\tabularnewline
\cline{1-2} \cline{7-14} \cline{17-24} \cline{27-35} 
\multicolumn{2}{c}{Set \# } & \multicolumn{2}{c}{$\vartheta$} & \multicolumn{2}{c}{~~} & \multicolumn{2}{c}{$\vartheta_{ref}$} & \multicolumn{2}{c}{$b_{BS}$} & \multicolumn{2}{c}{$I_{BS}$ } & \multicolumn{2}{c}{$\zeta_{BS}$ } & \multicolumn{2}{c}{~~} & \multicolumn{2}{c}{$\tilde{\vartheta}_{ref}$} & $u(\tilde{\vartheta}_{ref})$ & \multicolumn{2}{c}{$\zeta_{Sim}$} & \multicolumn{2}{c}{$I_{Sim2}$} & $\zeta_{Sim2}$ & \multicolumn{2}{c}{~~} & \multicolumn{2}{c}{$\tilde{\vartheta}_{ref}$} & \multicolumn{2}{c}{$u(\tilde{\vartheta}_{ref})$} & \multicolumn{2}{c}{$\zeta_{Sim}$} & \multicolumn{2}{c}{$I_{Sim2}$} & $\zeta_{Sim2}$\tabularnewline
\cline{1-4} \cline{7-14} \cline{17-24} \cline{27-35} 
\multicolumn{2}{c}{1 } & 0&96 & \multicolumn{2}{c}{} & \multicolumn{2}{c}{1} & 5&3e-04 & {[}0&87, 1.12{]} & -0&25 & \multicolumn{2}{c}{} & 1&00  & 3.1e-04 & -0&25  & {[}0&94, 1.06{]} & -0.66 & \multicolumn{2}{c}{} & 1&00 & 5&0e-04 & -0&26 & {[}0&91, 1.10{]} & -0.45\tabularnewline
\multicolumn{2}{c}{2 } & 0&86 & \multicolumn{2}{c}{} & \multicolumn{2}{c}{1} & 2&6e-04 & {[}0&80, 0.99{]} & -1&09 & \multicolumn{2}{c}{} & 1&00  & 2.3e-04 & -1&09  & {[}0&96, 1.04{]} & -2.65 & \multicolumn{2}{c}{} & 1&00 & 3&7e-04 & -1&09 & {[}0&93, 1.07{]} & -1.72\tabularnewline
\multicolumn{2}{c}{3 } & 1&12 & \multicolumn{2}{c}{} & \multicolumn{2}{c}{1} & 6&5e-04 & {[}1&05, 1.20{]} & 1&66 & \multicolumn{2}{c}{} & 1&00  & 3.2e-04 & 1&66  & {[}0&94, 1.06{]} & 1.89 & \multicolumn{2}{c}{} & 1&00 & 4&9e-04 & 1&67 & {[}0&91, 1.10{]} & 1.17\tabularnewline
\multicolumn{2}{c}{4 } & 1&23 & \multicolumn{2}{c}{} & \multicolumn{2}{c}{1} & -2&6e-04 & {[}1&16, 1.30{]} & 3&53 & \multicolumn{2}{c}{} & 1&00  & 2.3e-04 & 3&54  & {[}0&96, 1.04{]} & 5.01 & \multicolumn{2}{c}{} & 1&00 & 3&6e-04 & 3&53 & {[}0&94, 1.08{]} & 3.00\tabularnewline
\multicolumn{2}{c}{5 } & 0&85 & \multicolumn{2}{c}{} & \multicolumn{2}{c}{1} & 5&8e-04 & {[}0&78, 0.92{]} & -1&99 & \multicolumn{2}{c}{} & 1&00  & 3.1e-04 & -1&99  & {[}0&94, 1.06{]} & -2.54 & \multicolumn{2}{c}{} & 1&00 & 4&9e-04 & -1&98 & {[}0&91, 1.10{]} & -1.73\tabularnewline
\multicolumn{2}{c}{6 } & 0&98 & \multicolumn{2}{c}{} & \multicolumn{2}{c}{1} & -9&1e-04 & {[}0&86, 1.15{]} & -0&10 & \multicolumn{2}{c}{} & 1&00  & 2.3e-04 & -0&09  & {[}0&95, 1.04{]} & -0.35 & \multicolumn{2}{c}{} & 1&00 & 3&6e-04 & -0&10 & {[}0&94, 1.07{]} & -0.25\tabularnewline
\multicolumn{2}{c}{7 } & 0&97 & \multicolumn{2}{c}{} & \multicolumn{2}{c}{1} & 4&9e-04 & {[}0&94, 1.01{]} & -0&71 & \multicolumn{2}{c}{} & 1&00  & 1.2e-04 & -0&71  & {[}0&98, 1.02{]} & -1.23 & \multicolumn{2}{c}{} & 1&00  & 1&9e-04 & -0&72 & {[}0&96, 1.04{]} & -0.78\tabularnewline
\multicolumn{2}{c}{8 } & 0&93 & \multicolumn{2}{c}{} & \multicolumn{2}{c}{1} & -2&9e-04 & {[}0&87, 0.99{]} & -1&16 & \multicolumn{2}{c}{} & 1&00  & 2.0e-04 & -1&16  & {[}0&96, 1.04{]} & -1.92 & \multicolumn{2}{c}{} & 1&00 & 3&1e-04 & -1&15 & {[}0&94, 1.06{]} & -1.27\tabularnewline
\multicolumn{2}{c}{9 } & 0&97 & \multicolumn{2}{c}{} & \multicolumn{2}{c}{1} & 1&5e-03 & {[}0&90, 1.08{]} & -0&27 & \multicolumn{2}{c}{} & 1&00  & 2.0e-04 & -0&27  & {[}0&96, 1.04{]} & -0.74 & \multicolumn{2}{c}{} & 1&00 & 3&2e-04 & -0&27 & {[}0&94, 1.06{]} & -0.50\tabularnewline
\hline 
\end{tabular}
\end{turn}{\footnotesize\par}
\par\end{centering}
{\footnotesize{}\caption{\label{tab:ZMS}Intermediate values for the estimation of $\zeta$-scores
for the ZMS statistics. }
}{\footnotesize\par}
\end{table}
 
\begin{table}[t]
\noindent \begin{centering}
{\footnotesize{}}%
\begin{turn}{90}
\begin{tabular}{r@{\extracolsep{0pt}.}lr@{\extracolsep{0pt}.}lr@{\extracolsep{0pt}.}lr@{\extracolsep{0pt}.}lr@{\extracolsep{0pt}.}lr@{\extracolsep{0pt}.}lr@{\extracolsep{0pt}.}lr@{\extracolsep{0pt}.}lr@{\extracolsep{0pt}.}lcr@{\extracolsep{0pt}.}lr@{\extracolsep{0pt}.}lcr@{\extracolsep{0pt}.}lr@{\extracolsep{0pt}.}lr@{\extracolsep{0pt}.}lr@{\extracolsep{0pt}.}lcc}
\hline 
\multicolumn{2}{|c|}{$\vartheta=CC$~} & \multicolumn{2}{c}{} & \multicolumn{2}{c}{} & \multicolumn{8}{c}{BS} & \multicolumn{2}{c}{} & \multicolumn{8}{c}{SimN, Sim2N} & \multicolumn{2}{c}{} & \multicolumn{8}{c}{SimT, Sim2T}\tabularnewline
\cline{1-2} \cline{7-14} \cline{17-24} \cline{27-34} 
\multicolumn{2}{c}{Set \# } & \multicolumn{2}{c}{$\vartheta$} & \multicolumn{2}{c}{~~} & \multicolumn{2}{c}{$\vartheta_{ref}$} & \multicolumn{2}{c}{$b_{BS}$} & \multicolumn{2}{c}{$I_{BS}$ } & \multicolumn{2}{c}{$\zeta_{BS}$ } & \multicolumn{2}{c}{~~} & \multicolumn{2}{c}{$\tilde{\vartheta}_{ref}$} & $u(\tilde{\vartheta}_{ref})$ & \multicolumn{2}{c}{$\zeta_{Sim}$} & \multicolumn{2}{c}{$I_{Sim2}$} & $\zeta_{Sim2}$ & \multicolumn{2}{c}{~~} & \multicolumn{2}{c}{$\tilde{\vartheta}_{ref}$} & \multicolumn{2}{c}{$u(\tilde{\vartheta}_{ref})$} & \multicolumn{2}{c}{$\zeta_{Sim}$} & $I_{Sim2}$ & $\zeta_{Sim2}$\tabularnewline
\cline{1-4} \cline{7-14} \cline{17-24} \cline{27-34} 
\multicolumn{2}{c}{1 } & 0&50 & \multicolumn{2}{c}{} & \multicolumn{2}{c}{n/a} & -3&1e-05 & {[}0&467, 0.536{]} & \multicolumn{2}{c}{n/a} & \multicolumn{2}{c}{} & 0&40 & 1.9e-04  & 2&76  & {[}0&38, 0.44{]} & 2.67  & \multicolumn{2}{c}{} & 0&38 & 1&9e-04  & 3&39  & {[}0.34, 0.42{]} & 3.35\tabularnewline
\multicolumn{2}{c}{2 } & 0&62 & \multicolumn{2}{c}{} & \multicolumn{2}{c}{n/a} & -1&2e-04 & {[}0&598, 0.641{]} & \multicolumn{2}{c}{n/a} & \multicolumn{2}{c}{} & 0&57 & 1.1e-04  & 2&40  & {[}0&55, 0.59{]} & 2.38  & \multicolumn{2}{c}{} & 0&55 & 1&1e-04  & 3&44  & {[}0.52, 0.57{]} & 3.37\tabularnewline
\multicolumn{2}{c}{3 } & 0&26 & \multicolumn{2}{c}{} & \multicolumn{2}{c}{n/a} & -1&8e-04 & {[}0&216, 0.300{]} & \multicolumn{2}{c}{n/a} & \multicolumn{2}{c}{} & 0&25 & 2.1e-04  & 0&21  & {[}0&21, 0.29{]} & 0.21  & \multicolumn{2}{c}{} & 0&23 & 2&1e-04  & 0&61  & {[}0.19, 0.27{]} & 0.60\tabularnewline
\multicolumn{2}{c}{4 } & 0&40 & \multicolumn{2}{c}{} & \multicolumn{2}{c}{n/a} & 6&6e-05 & {[}0&372, 0.428{]} & \multicolumn{2}{c}{n/a} & \multicolumn{2}{c}{} & 0&42 & 1.3e-04  & -0&77  & {[}0&40, 0.45{]} & -0.78  & \multicolumn{2}{c}{} & 0&40 & 1&4e-04  & 0&05  & {[}0.37, 0.43{]} & 0.05\tabularnewline
\multicolumn{2}{c}{5 } & 0&04 & \multicolumn{2}{c}{} & \multicolumn{2}{c}{n/a} & -1&5e-04 & {[}-0&004, 0.081{]} & \multicolumn{2}{c}{n/a} & \multicolumn{2}{c}{} & 0&11 & 2.2e-04  & -1&63  & {[}0&07, 0.15{]} & -1.65  & \multicolumn{2}{c}{} & 0&10 & 2&2e-04  & -1&45  & {[}0.06, 0.14{]} & -1.47\tabularnewline
\multicolumn{2}{c}{6 } & 0&40 & \multicolumn{2}{c}{} & \multicolumn{2}{c}{n/a} & -1&4e-04 & {[}0&373, 0.433{]} & \multicolumn{2}{c}{n/a} & \multicolumn{2}{c}{} & 0&50 & 1.2e-04  & -3&27  & {[}0&48, 0.52{]} & -4.10  & \multicolumn{2}{c}{} & 0&48 & 1&2e-04  & -2&57  & {[}0.46, 0.50{]} & -3.26\tabularnewline
\multicolumn{2}{c}{7 } & 0&31 & \multicolumn{2}{c}{} & \multicolumn{2}{c}{n/a} & 7&9e-06 & {[}0&297, 0.328{]} & \multicolumn{2}{c}{n/a} & \multicolumn{2}{c}{} & 0&37 & 7.2e-05  & -3&86  & {[}0&36, 0.39{]} & -4.27  & \multicolumn{2}{c}{} & 0&35 & 7&5e-05  & -2&54  & {[}0.34, 0.37{]} & -2.67\tabularnewline
\multicolumn{2}{c}{8 } & -0&03 & \multicolumn{2}{c}{} & \multicolumn{2}{c}{n/a} & -1&2e-05 & {[}-0&052, 0.003{]} & \multicolumn{2}{c}{n/a} & \multicolumn{2}{c}{} & 0&11 & 1.4e-04  & -4&92  & {[}0&08, 0.14{]} & -5.06  & \multicolumn{2}{c}{} & 0&10 & 1&4e-04  & -4&61  & {[}0.08, 0.13{]} & -4.62\tabularnewline
\multicolumn{2}{c}{9 } & 0&23 & \multicolumn{2}{c}{} & \multicolumn{2}{c}{n/a} & -1&0e-04 & {[}0&207, 0.258{]} & \multicolumn{2}{c}{n/a} & \multicolumn{2}{c}{} & 0&13 & 1.4e-04  & 3&82  & {[}0&10, 0.16{]} & 3.77  & \multicolumn{2}{c}{} & 0&12 & 1&4e-04  & 4&19  & {[}0.09, 0.15{]} & 4.17\tabularnewline
\hline 
\end{tabular}
\end{turn}{\footnotesize\par}
\par\end{centering}
\caption{\label{tab:CC}Same as Table\,\ref{tab:ZMS} for the CC statistic. }
\end{table}

\noindent 
\begin{table}[t]
\noindent \begin{centering}
{\footnotesize{}}%
\begin{turn}{90}
\begin{tabular}{r@{\extracolsep{0pt}.}lr@{\extracolsep{0pt}.}lr@{\extracolsep{0pt}.}lr@{\extracolsep{0pt}.}lr@{\extracolsep{0pt}.}lr@{\extracolsep{0pt}.}lr@{\extracolsep{0pt}.}lr@{\extracolsep{0pt}.}lr@{\extracolsep{0pt}.}lcr@{\extracolsep{0pt}.}lr@{\extracolsep{0pt}.}lcr@{\extracolsep{0pt}.}lr@{\extracolsep{0pt}.}lr@{\extracolsep{0pt}.}lr@{\extracolsep{0pt}.}lcc}
\hline 
\multicolumn{2}{|c|}{$\vartheta=ENCE$~} & \multicolumn{2}{c}{} & \multicolumn{2}{c}{} & \multicolumn{8}{c}{BS} & \multicolumn{2}{c}{} & \multicolumn{8}{c}{SimN, Sim2N} & \multicolumn{2}{c}{} & \multicolumn{8}{c}{SimT, Sim2T}\tabularnewline
\cline{1-2} \cline{7-14} \cline{17-24} \cline{27-34} 
\multicolumn{2}{c}{Set \# } & \multicolumn{2}{c}{$\vartheta$} & \multicolumn{2}{c}{~~} & \multicolumn{2}{c}{$\vartheta_{ref}$} & \multicolumn{2}{c}{$b_{BS}$} & \multicolumn{2}{c}{$I_{BS}$ } & \multicolumn{2}{c}{$\zeta_{BS}$ } & \multicolumn{2}{c}{~~} & \multicolumn{2}{c}{$\tilde{\vartheta}_{ref}$} & $u(\tilde{\vartheta}_{ref})$ & \multicolumn{2}{c}{$\zeta_{Sim}$} & \multicolumn{2}{c}{$I_{Sim2}$} & $\zeta_{Sim2}$ & \multicolumn{2}{c}{~~} & \multicolumn{2}{c}{$\tilde{\vartheta}_{ref}$} & \multicolumn{2}{c}{$u(\tilde{\vartheta}_{ref})$} & \multicolumn{2}{c}{$\zeta_{Sim}$} & $I_{Sim2}$ & $\zeta_{Sim2}$\tabularnewline
\cline{1-4} \cline{7-14} \cline{17-24} \cline{27-34} 
\multicolumn{2}{c}{1 } & 0&125  & \multicolumn{2}{c}{} & \multicolumn{2}{c}{n/a} & 0&016  & {[}0&084, 0.153{]}  & \multicolumn{2}{c}{n/a} & \multicolumn{2}{c}{} & 0&056 & 9.5e-05  & 1&66  & {[}0&038, 0.076{]} & 3.52  & \multicolumn{2}{c}{} & 0&082  & 1&5e-04  & 1&04  & {[}0.055, 0.114{]} & 1.36\tabularnewline
\multicolumn{2}{c}{2 } & 0&126  & \multicolumn{2}{c}{} & \multicolumn{2}{c}{n/a} & 0&033  & {[}0&096, 0.130{]}  & \multicolumn{2}{c}{n/a} & \multicolumn{2}{c}{} & 0&041 & 7.0e-05  & 2&78  & {[}0&028, 0.056{]} & 5.98  & \multicolumn{2}{c}{} & 0&061  & 1&1e-04  & 2&13  & {[}0.041, 0.085{]} & 2.80\tabularnewline
\multicolumn{2}{c}{3 } & 0&097  & \multicolumn{2}{c}{} & \multicolumn{2}{c}{n/a} & 0&024  & {[}0&074, 0.101{]}  & \multicolumn{2}{c}{n/a} & \multicolumn{2}{c}{} & 0&058 & 9.8e-05  & 1&70  & {[}0&040, 0.077{]} & 1.98  & \multicolumn{2}{c}{} & 0&083  & 1&5e-04  & 0&59  & {[}0.056, 0.115{]} & 0.44\tabularnewline
\multicolumn{2}{c}{4 } & 0&135  & \multicolumn{2}{c}{} & \multicolumn{2}{c}{n/a} & 0&007  & {[}0&103, 0.157{]}  & \multicolumn{2}{c}{n/a} & \multicolumn{2}{c}{} & 0&043 & 7.4e-05  & 2&94  & {[}0&029, 0.058{]} & 5.88  & \multicolumn{2}{c}{} & 0&063  & 1&2e-04  & 2&30  & {[}0.043, 0.088{]} & 2.88\tabularnewline
\multicolumn{2}{c}{5 } & 0&131  & \multicolumn{2}{c}{} & \multicolumn{2}{c}{n/a} & 0&023  & {[}0&101, 0.139{]}  & \multicolumn{2}{c}{n/a} & \multicolumn{2}{c}{} & 0&056 & 9.4e-05  & 2&52  & {[}0&039, 0.075{]} & 3.88  & \multicolumn{2}{c}{} & 0&082  & 1&5e-04  & 1&66  & {[}0.055, 0.114{]} & 1.52\tabularnewline
\multicolumn{2}{c}{6 } & 0&244  & \multicolumn{2}{c}{} & \multicolumn{2}{c}{n/a} & 0&044  & {[}0&156, 0.276{]}  & \multicolumn{2}{c}{n/a} & \multicolumn{2}{c}{} & 0&045 & 8.1e-05  & 2&26  & {[}0&030, 0.062{]} & 11.7  & \multicolumn{2}{c}{} & 0&066  & 1&3e-04  & 2&02  & {[}0.044, 0.095{]} & 6.21\tabularnewline
\multicolumn{2}{c}{7 } & 0&066  & \multicolumn{2}{c}{} & \multicolumn{2}{c}{n/a} & 0&006  & {[}0&045, 0.085{]}  & \multicolumn{2}{c}{n/a} & \multicolumn{2}{c}{} & 0&026 & 5.3e-05  & 1&93  & {[}0&017, 0.037{]} & 3.58  & \multicolumn{2}{c}{} & 0&038  & 8&3e-05  & 1&33  & {[}0.025, 0.056{]} & 1.55\tabularnewline
\multicolumn{2}{c}{8 } & 0&108  & \multicolumn{2}{c}{} & \multicolumn{2}{c}{n/a} & 0&015  & {[}0&077, 0.118{]}  & \multicolumn{2}{c}{n/a} & \multicolumn{2}{c}{} & 0&036 & 6.0e-05  & 2&28  & {[}0&025, 0.048{]} & 5.77  & \multicolumn{2}{c}{} & 0&053  & 9&7e-05  & 1&72  & {[}0.036, 0.074{]} & 2.65\tabularnewline
\multicolumn{2}{c}{9 } & 0&120  & \multicolumn{2}{c}{} & \multicolumn{2}{c}{n/a} & 0&013  & {[}0&082, 0.140{]}  & \multicolumn{2}{c}{n/a} & \multicolumn{2}{c}{} & 0&036 & 6.0e-05  & 2&21  & {[}0&025, 0.048{]} & 6.95  & \multicolumn{2}{c}{} & 0&054  & 9&7e-05  & 1&74  & {[}0.036, 0.074{]} & 3.25\tabularnewline
\hline 
\end{tabular}
\end{turn}{\footnotesize\par}
\par\end{centering}
\caption{\label{tab:ENCE}Same as Table\,\ref{tab:ZMS} for the ENCE statistic. }
\end{table}

\noindent 
\begin{table}[t]
\noindent \begin{centering}
{\footnotesize{}}%
\begin{turn}{90}
\begin{tabular}{r@{\extracolsep{0pt}.}lr@{\extracolsep{0pt}.}lr@{\extracolsep{0pt}.}lr@{\extracolsep{0pt}.}lr@{\extracolsep{0pt}.}lr@{\extracolsep{0pt}.}lr@{\extracolsep{0pt}.}lr@{\extracolsep{0pt}.}lr@{\extracolsep{0pt}.}lcr@{\extracolsep{0pt}.}lr@{\extracolsep{0pt}.}lcr@{\extracolsep{0pt}.}lr@{\extracolsep{0pt}.}lr@{\extracolsep{0pt}.}lr@{\extracolsep{0pt}.}lcc}
\hline 
\multicolumn{2}{|c|}{$\vartheta=ZMSE$~} & \multicolumn{2}{c}{} & \multicolumn{2}{c}{} & \multicolumn{8}{c}{BS} & \multicolumn{2}{c}{} & \multicolumn{8}{c}{SimN, Sim2N} & \multicolumn{2}{c}{} & \multicolumn{8}{c}{SimT, Sim2T}\tabularnewline
\cline{1-2} \cline{7-14} \cline{17-24} \cline{27-34} 
\multicolumn{2}{c}{Set \# } & \multicolumn{2}{c}{$\vartheta$} & \multicolumn{2}{c}{~~} & \multicolumn{2}{c}{$\vartheta_{ref}$} & \multicolumn{2}{c}{$b_{BS}$} & \multicolumn{2}{c}{$I_{BS}$ } & \multicolumn{2}{c}{$\zeta_{BS}$ } & \multicolumn{2}{c}{~~} & \multicolumn{2}{c}{$\tilde{\vartheta}_{ref}$} & $u(\tilde{\vartheta}_{ref})$ & \multicolumn{2}{c}{$\zeta_{Sim}$} & \multicolumn{2}{c}{$I_{Sim2}$} & $\zeta_{Sim2}$ & \multicolumn{2}{c}{~~} & \multicolumn{2}{c}{$\tilde{\vartheta}_{ref}$} & \multicolumn{2}{c}{$u(\tilde{\vartheta}_{ref})$} & \multicolumn{2}{c}{$\zeta_{Sim}$} & $I_{Sim2}$ & $\zeta_{Sim2}$\tabularnewline
\cline{1-4} \cline{7-14} \cline{17-24} \cline{27-34} 
\multicolumn{2}{c}{1 } & 0&255 & \multicolumn{2}{c}{} & \multicolumn{2}{c}{n/a} & 0&034 & {[}0&172, 0.299{]} & \multicolumn{2}{c}{n/a} & \multicolumn{2}{c}{} & 0&112 & 1.9e-04 & 1&71 & {[}0&077, 0.152{]} & 3.63 & \multicolumn{2}{c}{} & 0&164 & 2&9e-04 & 1&09 & {[}0.111, 0.223{]} & 1.54\tabularnewline
\multicolumn{2}{c}{2 } & 0&273 & \multicolumn{2}{c}{} & \multicolumn{2}{c}{n/a} & 0&072 & {[}0&207, 0.283{]} & \multicolumn{2}{c}{n/a} & \multicolumn{2}{c}{} & 0&082 & 1.4e-04 & 2&91 & {[}0&056, 0.111{]} & 6.62 & \multicolumn{2}{c}{} & 0&121 & 2&1e-04 & 2&31 & {[}0.083, 0.165{]} & 3.46\tabularnewline
\multicolumn{2}{c}{3 } & 0&173 & \multicolumn{2}{c}{} & \multicolumn{2}{c}{n/a} & 0&048 & {[}0&136, 0.180{]} & \multicolumn{2}{c}{n/a} & \multicolumn{2}{c}{} & 0&112 & 1.9e-04 & 1&64 & {[}0&077, 0.151{]} & 1.57 & \multicolumn{2}{c}{} & 0&163 & 2&8e-04 & 0&26 & {[}0.110, 0.223{]} & 0.16\tabularnewline
\multicolumn{2}{c}{4 } & 0&247 & \multicolumn{2}{c}{} & \multicolumn{2}{c}{n/a} & 0&011 & {[}0&191, 0.287{]} & \multicolumn{2}{c}{n/a} & \multicolumn{2}{c}{} & 0&082 & 1.4e-04 & 2&97 & {[}0&056, 0.110{]} & 5.81 & \multicolumn{2}{c}{} & 0&121 & 2&1e-04 & 2&26 & {[}0.083, 0.165{]} & 2.87\tabularnewline
\multicolumn{2}{c}{5 } & 0&283 & \multicolumn{2}{c}{} & \multicolumn{2}{c}{n/a} & 0&052 & {[}0&221, 0.304{]} & \multicolumn{2}{c}{n/a} & \multicolumn{2}{c}{} & 0&112 & 1.9e-04 & 2&77 & {[}0&077, 0.151{]} & 4.42 & \multicolumn{2}{c}{} & 0&163 & 2&9e-04 & 1&94 & {[}0.111, 0.224{]} & 1.96\tabularnewline
\multicolumn{2}{c}{6 } & 0&356 & \multicolumn{2}{c}{} & \multicolumn{2}{c}{n/a} & 0&270 & {[}0&240, 0.357{]} & \multicolumn{2}{c}{n/a} & \multicolumn{2}{c}{} & 0&082 & 1.4e-04 & 2&37 & {[}0&056, 0.111{]} & 9.53 & \multicolumn{2}{c}{} & 0&121 & 2&2e-04 & 2&02 & {[}0.082, 0.168{]} & 5.06\tabularnewline
\multicolumn{2}{c}{7 } & 0&118 & \multicolumn{2}{c}{} & \multicolumn{2}{c}{n/a} & 0&018 & {[}0&078, 0.131{]} & \multicolumn{2}{c}{n/a} & \multicolumn{2}{c}{} & 0&043 & 7.2e-05 & 1&87 & {[}0&029, 0.058{]} & 5.07 & \multicolumn{2}{c}{} & 0&066 & 1&2e-04 & 1&31 & {[}0.045, 0.089{]} & 2.20\tabularnewline
\multicolumn{2}{c}{8 } & 0&225 & \multicolumn{2}{c}{} & \multicolumn{2}{c}{n/a} & 0&033 & {[}0&162, 0.246{]} & \multicolumn{2}{c}{n/a} & \multicolumn{2}{c}{} & 0&071 & 1.2e-04 & 2&43 & {[}0&049, 0.097{]} & 6.08 & \multicolumn{2}{c}{} & 0&107 & 1&9e-04 & 1&87 & {[}0.072, 0.147{]} & 2.96\tabularnewline
\multicolumn{2}{c}{9 } & 0&250 & \multicolumn{2}{c}{} & \multicolumn{2}{c}{n/a} & 0&025 & {[}0&171, 0.287{]} & \multicolumn{2}{c}{n/a} & \multicolumn{2}{c}{} & 0&071 & 1.2e-04 & 2&27 & {[}0&049, 0.096{]} & 7.32 & \multicolumn{2}{c}{} & 0&107 & 1&9e-04 & 1&81 & {[}0.073, 0.147{]} & 3.62\tabularnewline
\hline 
\end{tabular}
\end{turn}{\footnotesize\par}
\par\end{centering}
\caption{\label{tab:ZMSE}Same as Table\,\ref{tab:ZMS} for the ZMSE statistic. }
\end{table}

All standardized $\zeta$-scores are built from the equation
\begin{equation}
\zeta(\vartheta_{est},\vartheta_{ref},I)=\begin{cases}
\frac{\vartheta_{est}-\vartheta_{ref}}{I^{+}-\vartheta_{est}} & if\,(\vartheta_{est}-\vartheta_{ref})\le0\\
\frac{\vartheta_{est}-\vartheta_{ref}}{\vartheta_{est}-I^{-}} & if\,(\vartheta_{est}-\vartheta_{ref})>0
\end{cases}\label{eq:zeta-def-1}
\end{equation}
where $\vartheta_{est}$ is the estimated value of the statistic,
$\vartheta_{ref}$ is its reference value, and $I=[I^{-},I^{+}]$
is a 95\% confidence interval. The values of $\vartheta_{ref}$ and
$I$ are estimated according to three schemes\\
~
\begin{center}
\begin{tabular}{lcc}
\hline 
Scheme & $\vartheta_{ref}$ & $I$\tabularnewline
\hline 
BS & Predefined & Bootstrapped\tabularnewline
Sim & Monte Carlo & Bootstrapped\tabularnewline
Sim2 & Monte Carlo & Monte Carlo\tabularnewline
\hline 
\end{tabular}\\
~
\par\end{center}

\noindent In the absence of a predefined reference value, $\zeta_{BS}$
cannot be estimated. 

In the Sim and Sim2 schemes, synthetic error sets are generated according
to
\begin{equation}
E_{i}=u_{E,i}\epsilon_{i}
\end{equation}
where $u_{E,i}$ is one of the uncertainties in the original dataset,
and $\epsilon_{i}$ is a random value from the generative distribution
$D$ ($\epsilon_{i}\sim D(0,1)$ ). Two variants of $D$ are considered
in the present study, denoted N and T:
\begin{description}
\item [{N}] a standard normal distribution $D=N(0,1)$ 
\item [{T}] a unit\_variance Student's-\emph{t} distribution $D=t_{s}(6)$
\end{description}
One has thus five options for the estimation of standardized scores:
$\zeta_{BS}$, $\zeta_{SimN}$, $\zeta_{Sim2N}$, $\zeta_{SimT}$,
$\zeta_{Sim2T}$, which are reported in the results tables with intermediate
results and in Fig.\,\ref{fig:z-Scores}.

\clearpage{}

\section{Reference values for the ENCE and ZMSE statistic\label{sec:Reference-values-for}}

The ENCE and ZMSE statistics have been shown previously to depend
on the binning scheme\citep{Pernot2023a_arXiv}. We consider here
other sources of influence, such as the dataset size, and uncertainty
distribution shape. This study is performed on synthetic datasets,
and the dependence on the generative model is also assessed.

\begin{figure}[t]
\noindent \begin{centering}
\includegraphics[width=0.75\textwidth]{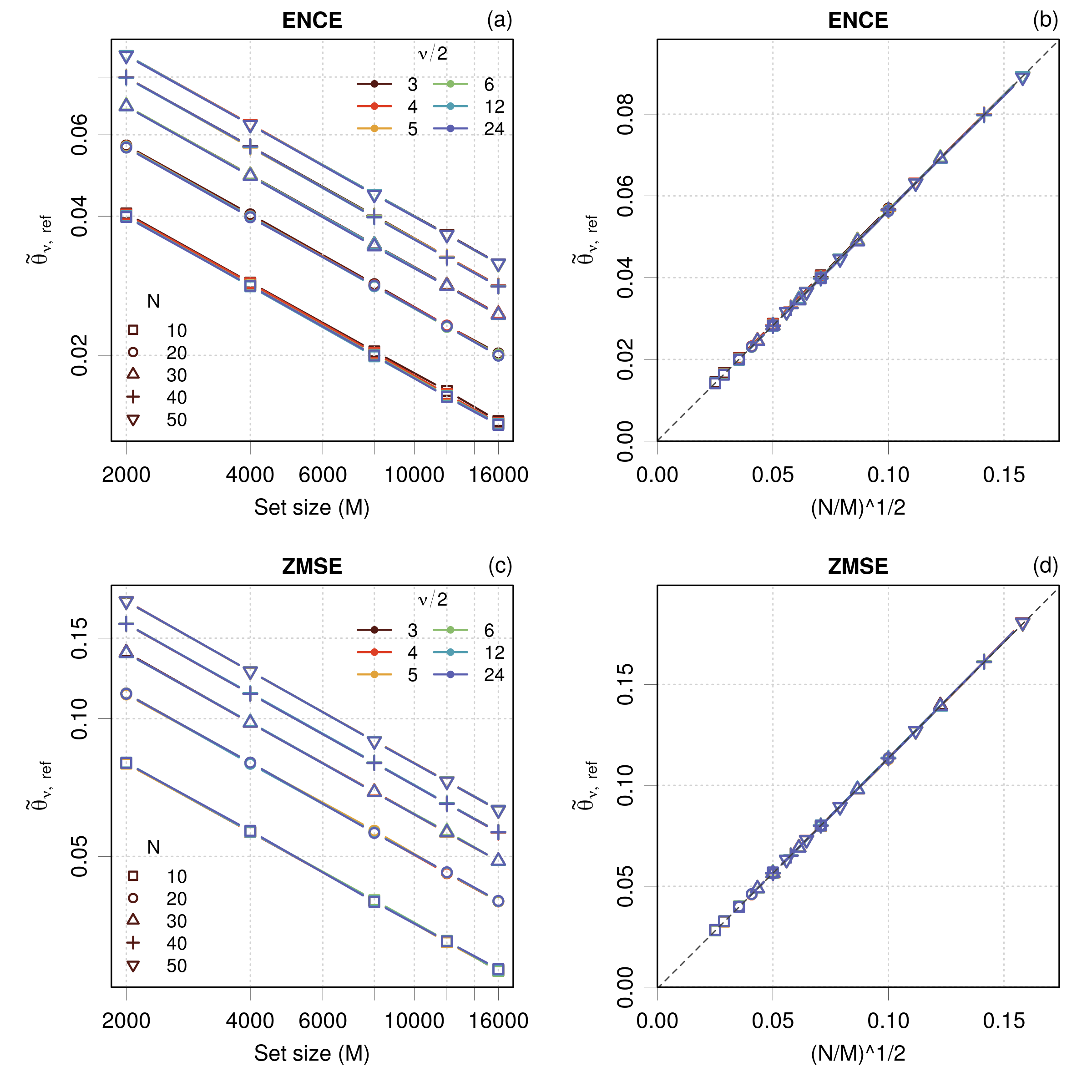}
\par\end{centering}
\caption{\label{fig:ENCE-ZMSE}Sensitivity of simulated ENCE and ZMSE values
to the dataset size ($M$), number of bins ($N$) and shape of the
uncertainty distribution ($\nu$), for calibrated datasets generated
from the \emph{NIG} model.}
\end{figure}

Simulated reference values $\tilde{\vartheta}_{D,ref}$ for the ENCE
and ZMSE statistics have been estimated from $N_{MC}=5000$ synthetic
calibrated datasets generated by the NIG model
\begin{align}
E & \sim u_{E}\times N(0,1)\\
u_{E}^{2} & \sim\Gamma^{-1}(\nu/2,\nu/2)
\end{align}
with three varying parameters: the dataset size $M\in\left\{ 2000,4000,8000,12000,16000\right\} $
, the number of equal-size bins $N\in\left\{ 10,20,30,40,50\right\} $
and the shape of the uncertainty distribution $\nu\in\left\{ 3,4,5,6,12,24\right\} $.
The relative uncertainties on the reported mean reference values are
around $1.5\thinspace10^{-3}$. The results are shown in Fig.\,\ref{fig:ENCE-ZMSE}.

The dependence of the ENCE on the number of bins was shown in a previous
study\citep{Pernot2023a_arXiv} to be in $N^{1/2}$. The log-log plot
in Fig.\,\ref{fig:ENCE-ZMSE}(left) reveals that the ENCE depends
also on $M$ as a power law. One has thus $ENCE=\alpha M^{\beta}N^{1/2}$
with $\beta\simeq-1/2$. The same data are then plot on a linear scale
as a function of $(N/M)^{1/2}$, from which one can see a perfect
linear fit by $\tilde{\vartheta}_{N,ref}=\alpha\times(N/M)^{1/2}$
with $\alpha\simeq0.56$. The ZMSE presents features very similar
to the ENCE, with a different slope, $\tilde{\vartheta}_{N,ref}\simeq1.14\times(N/M)^{1/2}$.
In both cases, the impact of the uncertainty distribution is minor,
albeit larger for ENCE than for ZMSE, and negligible for the validation
process. 

It seems therefore that, for a normal generative model and a chosen
binning scheme, it is possible to define a reference value for the
ENCE and ZMSE statistics for each dataset. The availability of a predefined
reference value would considerably simplify the validation approach,
when compared to the extrapolation-based one proposed by Pernot\citep{Pernot2023a_arXiv}.

\begin{figure}[t]
\noindent \begin{centering}
\includegraphics[width=0.75\textwidth]{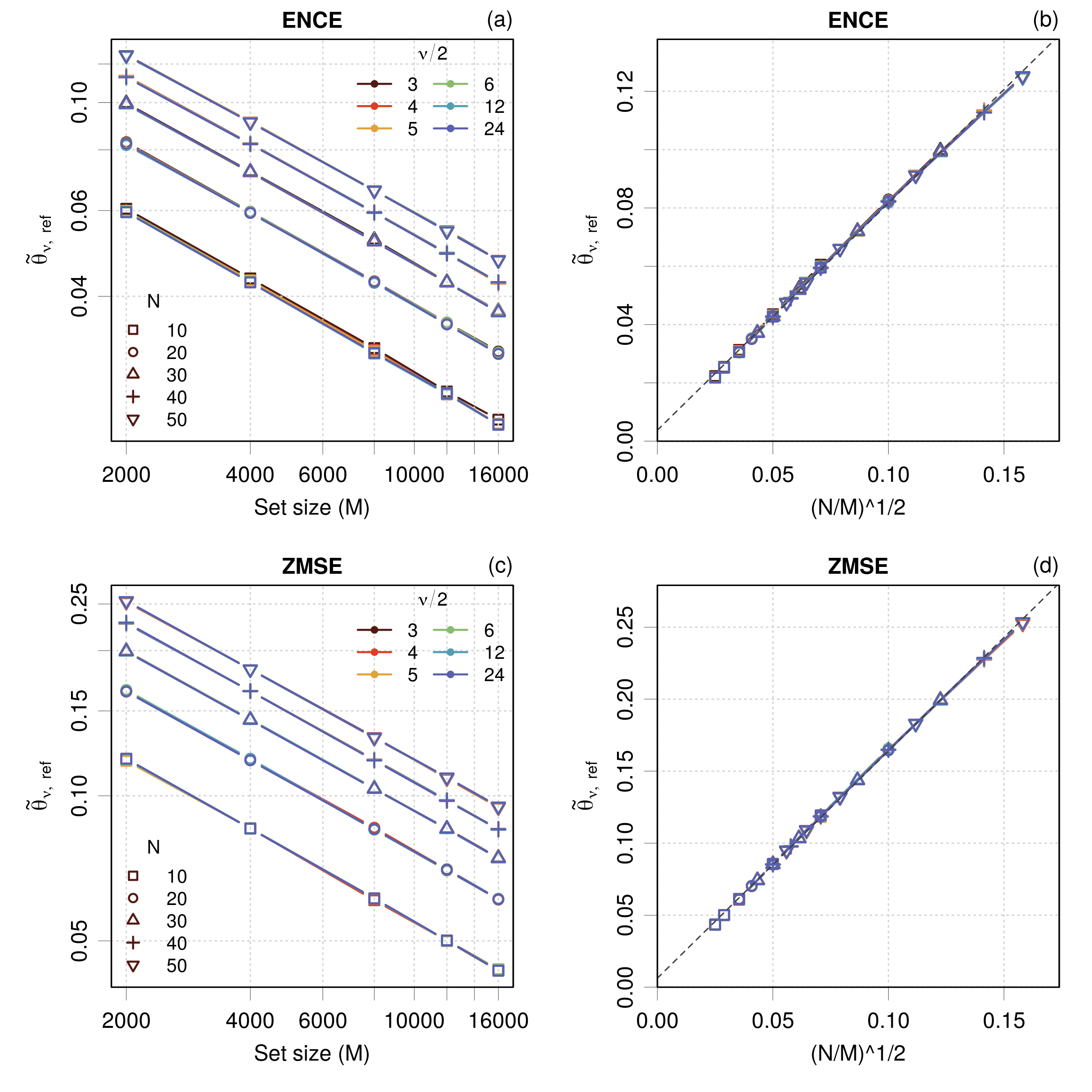}
\par\end{centering}
\caption{\label{fig:ENCE-ZMSE-1}Same as Fig.\,\ref{fig:ENCE-ZMSE} for the
T6IG model.}
\end{figure}

Let us now check the sensitivity of this approach to the generative
distribution. The same protocol as above is followed for another choice
of generative model (denoted T6IG)
\begin{align}
E & \sim u_{E}\times t_{s}(6)\\
u_{E}^{2} & \sim\Gamma^{-1}(\nu/2,\nu/2)
\end{align}
and the results are shown in Fig.\,\ref{fig:ENCE-ZMSE-1}.

The same linear relationship as for the normal generative distribution
is observed, but, in each case, with a different slope and a small
positive intercept. Besides, a very small deviation from linearity
can also be perceived. For ENCE, one gets then $\tilde{\vartheta}_{T,ref}\simeq0.004+0.779\times(N/M)^{1/2}$
and for ZMSE $\tilde{\vartheta}_{T,ref}\simeq0.006+1.577\times(N/M)^{1/2}$. 

It seems thus impossible to define a reference value for ENCE and
ZMSE if the generative distribution is unknown. 

\clearpage{}

\section{Validation of ZMSE for the application datasets\label{sec:Validation-of-ZMSE}}

For each of the nine datasets, one establishes a ZMSE validation diagnostic
by a method which is independent of a generative distribution hypothesis.
For this, the ZMSE is estimated for a series of bin numbers $N$ between
10 and 150, with the constraint that bins should not contain less
than 20 data points. The data with $N>20$ are fitted by a linear
model as a function of $(N/M)^{1/2}$. The intercept of the regression
line is then compared to zero.\citep{Pernot2023a_arXiv} The reference
lines for the NIG and T6IG models defined in Appendix\,\ref{sec:Reference-values-for}
are also plotted for comparison. 
\begin{figure}[t]
\noindent \begin{centering}
\includegraphics[width=0.95\textwidth]{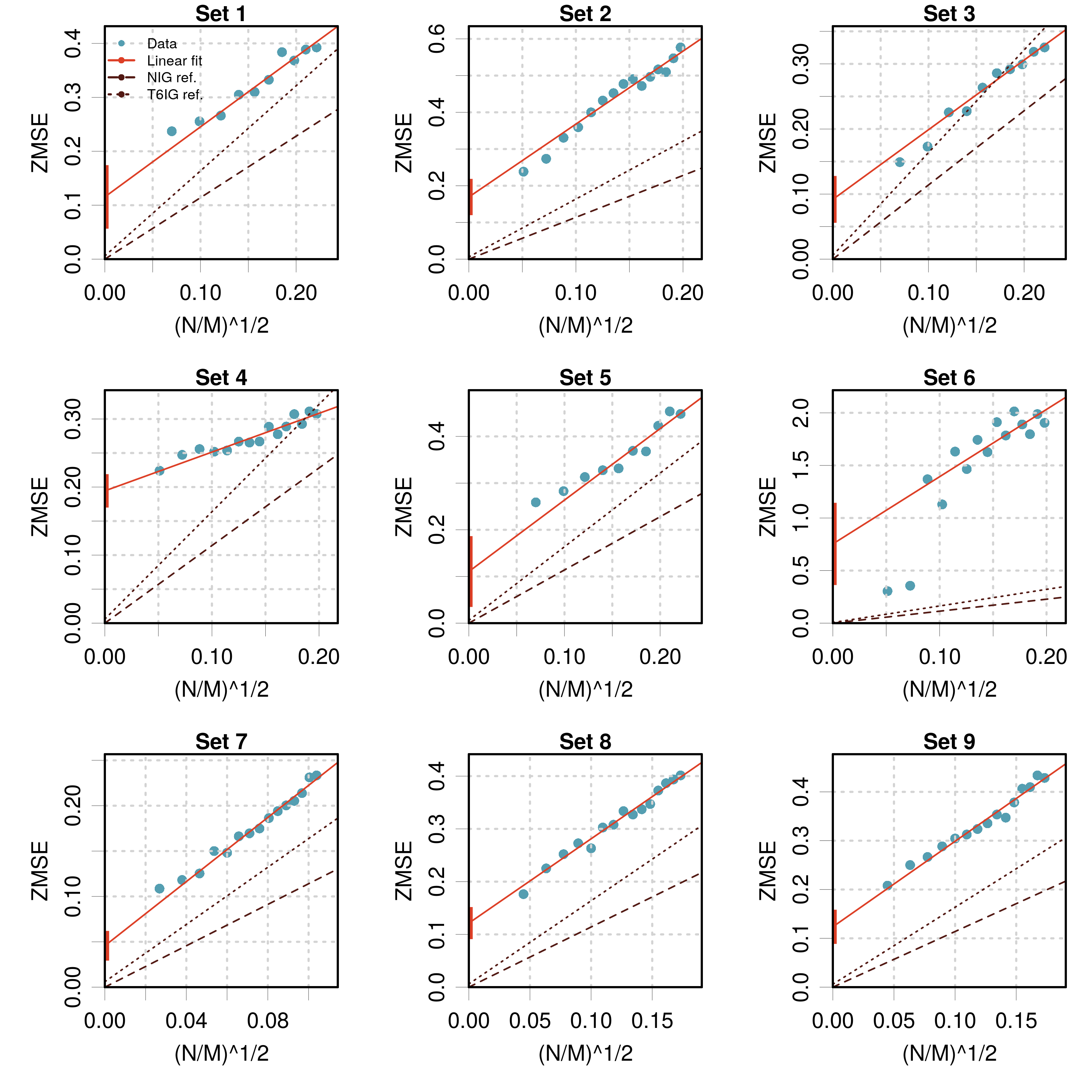}
\par\end{centering}
\caption{\label{figZMSEval} Validation of ZMSE by the extrapolation method:
ZMSE are estimated for a sequence of bin numbers and plotted as function
of $(N/M)^{1/2}$ (blue dots); this sequence (for $N>20$) is fitted
by a linear model (solid red line); reference lines are plotted for
the NIG (dashed line) and T6IG (dotted line) models.}
\end{figure}

All the datasets are failing this validation test of conditional calibration
(consistency), as none of the red intervals encloses the origin. Note
however that for some datasets (e.g. Sets 3 and 4), it is possible
to choose a number of bins such that the ZMSE is compatible with the
T6IG reference (dotted line). This shows the danger of using a reference
value from an unconstrained generative model.
\end{document}